\documentclass{article}

\PassOptionsToPackage{numbers, compress}{natbib}

\usepackage[eandd, final]{neurips_2026}

\usepackage[utf8]{inputenc} 
\usepackage[T1]{fontenc}    
\usepackage{hyperref}       
\usepackage{url}            
\usepackage{colortbl}
\usepackage{booktabs}       
\usepackage{amsfonts}       
\usepackage{nicefrac}       
\usepackage{microtype}      
\usepackage{xcolor}         
\usepackage{amsmath}
\usepackage{balance}
\usepackage{graphicx}
\usepackage{multirow} 
\usepackage{xspace}
\usepackage{wrapfig}
\newcommand{\ourq}{SRCoD\xspace}
\newcommand{\myparagraph}[1]{\vspace{1ex}\noindent\textbf{#1.}\hspace{1em}}

\author{%
\begin{tabular}{c}
Shiqi Zhang$^{1}$
\quad
Weixin Zeng$^{2}$\thanks{Corresponding author.}
\quad
Ziheng Zhang$^{2}$
\quad
Jiuyang Tang$^{2}$
\quad
Xiang Zhao$^{2}$
\\[0.6ex]
$^{1}$National Key Laboratory of Information Systems Engineering, \\
National University of Defense Technology, Changsha, China
\\[0.3ex]
$^{2}$National Key Laboratory of Big Data and Decision, \\
National University of Defense Technology, Changsha, China
\\[0.5ex]
\texttt{\{zhangshiqi,zengweixin13,zihengzhang20,jiuyang\_tang,xiangzhao\}@nudt.edu.cn}
\end{tabular}
}

\title{Are Multimodal Benchmarks Really Useful? Item-Level Multimodal Benchmark Diagnosis via Structure-Response Co-Calibration}

\begin{document}

\maketitle

\begin{abstract}
Multimodal benchmarks are central to evaluating whether models can reason over information from multiple modalities. However, existing literature has revealed that some benchmarks may contain shortcut items that can be answered from a single modality or from answer-option regularities alone, thus failing to reflect models' true cross-modal inference ability.
Existing methods for assessing benchmarks typically rely either on \textit{intrinsic item structure} or \textit{extrinsic model response behavior}, which are not adequate for a comprehensive diagnosis. 
To fill in this gap, we propose \ourq, a structure-response co-calibrated diagnosis framework for item-level diagnosis of modality dependence in multimodal benchmarks. \ourq estimates sample-wise multimodal information structure from benchmark content and target answers, and uses this signal as a soft structural anchor for modality-conditioned item response modeling. The learned model produces a calibrated item-level diagnostic profile that reflects how strongly an item depends on different forms of modality evidence, enabling fine-grained benchmark analysis beyond aggregate accuracy. 
\ourq also provides interpretable attribution of the diagnosis results.
Extensive experiments under two complementary protocols, i.e., benchmark-refinement oriented validation and human-labeled item validation, demonstrate that \ourq outperforms state-of-the-art benchmark diagnosis methods.    
Our code and data are available at \url{https://github.com/Q-17/SRCoD}.
\end{abstract}

\section{Introduction}

Multimodal benchmarks are central to evaluating the cross-modal reasoning ability of vision-language models (VLMs)~\citep{VLM1,VLM2,VLM3}. 
These benchmarks are in essence collections of \emph{items}, where each \emph{item} corresponds to a complete problem instance, usually comprising multimodal inputs, a question, and a ground-truth answer~\citep{ScienceQA,MMMU,SeedBench,MMBench,EMMA}.
Such benchmarks are designed to evaluate whether VLM can understand and reason across modalities~\cite{hima,comco}, and have inspired the development of state-of-the-art VLM models.
However, recently, the quality of these benchmarks, especially whether the items are all useful for evaluating models' cross-modal inference ability, has attracted increasing attention. 
Early efforts have revealed that existing multimodal benchmarks often contain shortcut items that can be answered from a single modality or from answer-option regularities alone~\citep{VQAv2,MMStar}. 
Therefore, \emph{item-level multimodal benchmark diagnosis}, which aims to identify what modality evidence each item actually requires and whether it provides a valid test of the intended cross-modal reasoning ability of downstream models, has become an important topic. 

Existing diagnostic methods can be categorized into two groups: methods based on \textit{intrinsic item structure} and methods based on \textit{extrinsic model responses}.
Intrinsic-structure based methods~\citep{PIDforEvaluation1,LSMI} characterize how answer-relevant evidence is distributed across modalities within the item itself, often through information-theoretic decomposition frameworks such as partial information decomposition (PID)~\citep{PID}. 
However, item structure alone cannot reflect how downstream models actually solve the item, as an item may contain cross-modal evidence that models would not use. For example, a sidewalk-color question may contain the necessary visual evidence, yet its question and answer options can strongly favor ``gray'' even without the image.
Extrinsic-response based methods, on the other hand, evaluate the quality of items from model behavior under different modality conditions, using item response theory (IRT)~\citep{irt}, multidimensional IRT~\citep{mirt}, or multimodal variants such as M3IRT~\citep{M3IRT}.
These methods first select a pool of downstream VLMs, collect their responses to each item under different modality conditions, and evaluate the multimodal benchmark from the resulting response matrix.
While this grounds the diagnosis in actual model behavior, it discards the intrinsic structure of items and makes the assessment solely based on the chosen model pool.
A narrow or imbalanced pool may yield response matrix that reflect model-specific strengths rather than the evidence an item actually requires. 
For example, identifying the resting-potential points in an action-potential graph requires both the textual question and the numbered graph. Yet failures by models with limited chart-understanding abilities may not reflect the item's intrinsic evidence requirements.

We therefore argue that reliable multimodal benchmark diagnosis requires \emph{a structure-response co-calibrated view}. 
Intrinsic item structure provides a structural signal about what evidence an item contains, while extrinsic model responses provide a behavioral signal about how the item functions in evaluation. 
We instantiate this view with \ourq, a \textbf{S}tructure-\textbf{R}esponse \textbf{Co}-calibrated \textbf{D}iagnosis framework for item-level multimodal benchmark diagnosis.
\ourq first derives a structural prior from sample-wise multimodal interaction signals, summarizing how answer-relevant evidence is distributed across modalities within the item. 
It then uses this prior to anchor a modality-conditioned IRT model fitted to extrinsic model responses under different modality settings. The resulting diagnostic profile therefore provides an attribution-style interpretation of item behavior, with each coordinate corresponding to one evidence factor.
In this way, \ourq preserves the intrinsic structure of the item while correcting it using empirical evaluation behavior.

To comprehensively assess item-level multimodal benchmark diagnosis baselines, we use two complementary validation protocols: \textit{benchmark-refinement oriented validation} and \textit{human-labeled item validation}. Following prior extrinsic-response work~\citep{M3IRT}, we conduct benchmark-refinement oriented validation using model-ranking preservation as the validation criterion. In addition, we introduce human-labeled item validation and construct a diagnostic validation set across ScienceQA~\citep{ScienceQA}, MMMU~\citep{MMMU}, and SEED-Bench~\citep{SeedBench}, which directly verifies whether each method recovers the dominant evidence factor underlying each item. Unlike prior intrinsic-structure work that uses human judgments mainly to assess estimator plausibility~\citep{LSMI}, our protocol collects controlled human labels indicating whether an item can be solved from the visual input alone, the textual input alone, either modality, or only their combination. 
Across both protocols, \ourq outperforms intrinsic-structure and extrinsic-response baselines, with ablations confirming the benefits of structure-response co-calibration.
In this paper, our contributions are threefold:
\begin{itemize}
    \item We propose a \emph{structure-response co-calibrated} view for item-level multimodal benchmark diagnosis, and instantiate it with \ourq, a structure-anchored multimodal IRT framework that combines intrinsic item structure with extrinsic model responses.

    \item \ourq provides interpretable attribution of the diagnosis results, indicating whether the multimodal features of an item are synergistic, redundant, visual-dominant or textual-dominant.
    
    \item We validate \ourq under two complementary protocols, i.e., \textit{benchmark-refinement oriented validation} and \textit{human-labeled item validation}. Across both protocols, \ourq outperforms state-of-the-art evaluation methods.    
\end{itemize}

\section{Related Work}
\label{sec:related_work}

\myparagraph{Intrinsic-structure methods}
Intrinsic-structure methods analyze how answer-relevant evidence is distributed across modalities within an item. 
A representative line of work uses Partial Information Decomposition (PID)~\citep{PID} to decompose multimodal information into redundant, unique, and synergistic components. 
Early PID-based analyses~\citep{PIDforEvaluation1,PIDforEvaluation2} typically operate at the dataset-distribution level, while recent sample-wise estimators such as LSMI~\citep{LSMI} make it possible to estimate item-level multimodal interaction structure. 
On real-world data, these methods often use human judgments to assess the plausibility of estimated interaction signals. 
However, such judgments are not controlled item-level evidence-sufficiency labels, and therefore do not directly validate whether the inferred modality-dependence profile of each benchmark item is correct.

\myparagraph{Extrinsic-response methods}
Extrinsic-response methods infer benchmark properties from model behavior. 
A common strategy is to evaluate models under different modality conditions and model the resulting response patterns with Item Response Theory (IRT)~\citep{irt} or multidimensional IRT (MIRT)~\citep{mirt}, treating models as subjects and benchmark examples as items. 
Recent multimodal extensions further decompose responses into visual, textual, and joint factors~\citep{M3IRT}. 
These methods are often validated through benchmark refinement protocols, such as selecting compact subsets that preserve model rankings or avoid semi-synthetic low-quality items. 
Such protocols demonstrate downstream validation utility, but they do not provide ground-truth item-level diagnosis labels for naturally occurring benchmark questions.

\myparagraph{Positioning of this work}
Our work addresses both the methodological and evaluation gaps above. 
\ourq connects intrinsic item structure and extrinsic model responses through structure-response co-calibration, using structural interaction signals as soft anchors for modality-conditioned IRT and calibrating them with observed model responses. 
We further introduce a controlled human validation protocol for assessing evidence sufficiency, which provides item-level diagnostic labels for directly validating modality-dependence.

\section{Method}
\label{sec:method}

We develop \ourq, a structure-anchored multimodal IRT framework that instantiates structure-response co-calibration for item-level multimodal benchmark diagnosis. 
The intrinsic item structure summarizes how answer-relevant evidence is distributed across modalities within the item itself, while the extrinsic model responses capture how a pool of VLMs behaves under different modality conditions.
The resulting modality-dependence profile summarizes four forms of item-level evidence dependence: redundant, visual-dominant, textual-dominant, and synergistic. 
\ourq first derives a structural prior from sample-wise multimodal interaction signals, and then uses this prior to anchor a modality-conditioned IRT model fitted to extrinsic model responses. 
The fitted model produces a response-calibrated modality-dependence profile, which serves as the final diagnostic representation of the item.
Since each coordinate corresponds to a distinct evidence factor, the profile also provides an interpretable attribution of item behavior.

\begin{figure}[t]
    \centering
    \includegraphics[width=0.85\linewidth]{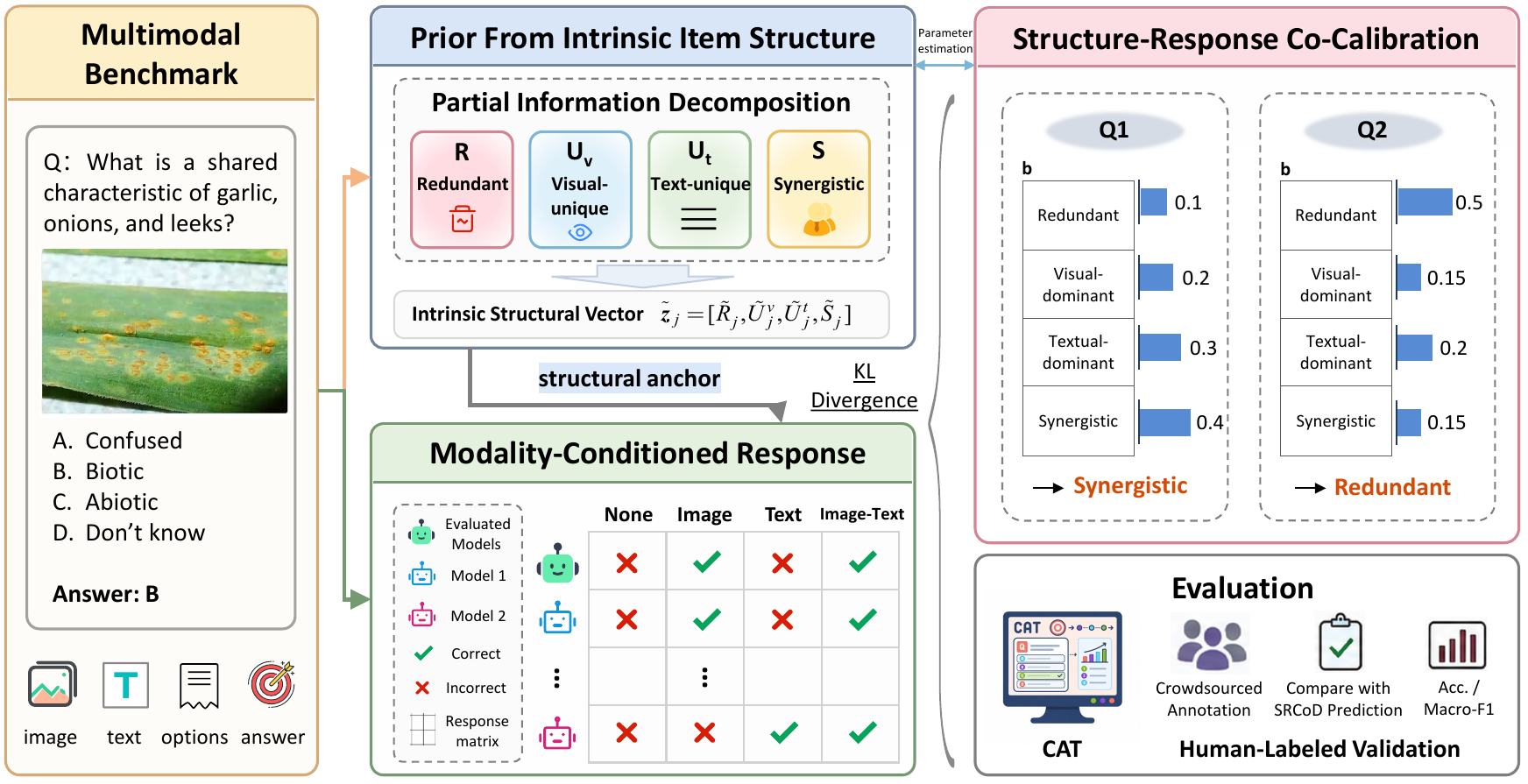}
    \caption{
    Overview of \ourq. 
    \ourq diagnoses item-level modality dependence by co-calibrating intrinsic item structure and extrinsic model responses. 
    It first derives a structural prior from partial information decomposition, then anchors a modality-conditioned response model fitted to VLM responses under controlled modality conditions.
    }
    \label{fig:SRCoD}
\end{figure}

\subsection{Modality-Conditioned Response Model}
\label{sec:modality_response_model}

Let \(\mathcal{D}=\{(v_j,t_j,y_j)\}_{j=1}^{J}\) denote a multimodal benchmark, where \(v_j\), \(t_j\), and \(y_j\) are the visual input, textual input, and ground-truth answer of item \(j\). 
For model \(i\), item \(j\), and modality condition \(c=(c_v,c_t)\in\{0,1\}^2\), we observe a binary response \(r_{ijc}\in\{0,1\}\), indicating whether the model answers correctly under condition \(c\). 

Following modality-conditioned multimodal IRT~\citep{M3IRT}, we use four response coordinates corresponding to a base coordinate, a visual coordinate, a textual coordinate, and a joint visual-textual coordinate. 
For condition \(c\), the active coordinates are encoded as
\(\boldsymbol{\phi}(c)=(1,c_v,c_t,c_vc_t)^\top\). 
Each model \(i\) has an ability vector \(\boldsymbol{\theta}_i\in\mathbb{R}^4\), and each item \(j\) has discrimination and difficulty vectors \(\mathbf{a}_j,\mathbf{b}_j\in\mathbb{R}^4\). 
The probability of a correct response is modeled as
\begin{equation}
    P_{ijc}
    =
    P(r_{ijc}=1)
    =
    \sigma\!\left(
    \boldsymbol{\phi}(c)^\top
    \left(
    \mathbf{a}_j\odot\boldsymbol{\theta}_i-\mathbf{b}_j
    \right)
    \right),
    \label{eq:response_model}
\end{equation}
where \(\odot\) denotes element-wise multiplication. 
This likelihood captures how model correctness changes when different modality coordinates are activated.

If \(\mathbf{a}_j\) and \(\mathbf{b}_j\) were learned solely from responses, the resulting diagnosis would be response-only and could be sensitive to the diagnostic model pool. 
The key distinction of \ourq is that these item parameters are anchored by intrinsic item structure before being calibrated by extrinsic model responses.

\subsection{Prior From Intrinsic Item Structure}
\label{sec:structural_prior}

Response-only diagnosis can conflate modality dependence with empirical difficulty. 
To reduce this confounding, \ourq introduces a structural prior derived from intrinsic item structure.

For each item \(j=(v_j,t_j,y_j)\), we estimate a sample-wise multimodal interaction vector using the Lightweight Sample-wise Multimodal Interaction (LSMI) estimator~\citep{LSMI}. 
After robust normalization, the interaction vector is written as
\begin{equation}
    \tilde{\mathbf{z}}_j
    =
    (\tilde{R}_j,\tilde{U}^{v}_j,\tilde{U}^{t}_j,\tilde{S}_j)^\top,
\end{equation}
where \(\tilde{R}_j\) denotes redundant evidence, \(\tilde{U}^{v}_j\) and \(\tilde{U}^{t}_j\) denote visual-unique and textual-unique evidence, and \(\tilde{S}_j\) denotes synergistic evidence.

We convert this interaction vector into structural dependence scores. 
First, we define modality answerability scores as
\(A_{j,v}=\tilde{R}_j+\tilde{U}^{v}_j\),
\(A_{j,t}=\tilde{R}_j+\tilde{U}^{t}_j\), and
\(A_{j,vt}=\tilde{R}_j+\tilde{U}^{v}_j+\tilde{U}^{t}_j+\tilde{S}_j\). 
The structural dependence scores are then given by
\begin{equation}
\boldsymbol{\psi}_j
=
\left(
\min(A_{j,v},A_{j,t}),
[A_{j,v}-A_{j,t}]_+,
[A_{j,t}-A_{j,v}]_+,
[A_{j,vt}-\max(A_{j,v},A_{j,t})]_+
\right),
\label{eq:structural_scores}
\end{equation}
where \([x]_+=\max(x,0)\). 
The four coordinates correspond to redundant, visual-dominant, textual-dominant, and synergistic dependence, respectively.

The structural prior is defined as
\begin{equation}
    \mathbf{q}^{\mathrm{str}}_j
    =
    \operatorname{softmax}(\boldsymbol{\psi}_j).
\end{equation}
Because intrinsic structural estimates can be ambiguous, \ourq does not treat this prior as a hard label. 
Instead, it assigns each item an entropy-based confidence weight,
\begin{equation}
    w_j
    =
    \left[
    1-
    \frac{H(\mathbf{q}^{\mathrm{str}}_j)}{\log 4}
    -\gamma
    \right]_+ .
\end{equation}
A near-uniform structural prior therefore has little influence, whereas a low-entropy prior contributes more strongly during structure-response co-calibration.

\subsection{Structure-Anchored Item Parameters}
\label{sec:structure_anchored_params}

The structural interaction coordinates and the response-side IRT coordinates are related but not identical. 
The structural coordinates describe how answer-relevant evidence is distributed within the item, whereas the IRT coordinates describe how model correctness changes under modality interventions. 
To connect these two views, \ourq learns a bridge from intrinsic item structure to modality-conditioned item parameters.

Let 
\begin{equation}
    \mathbf{z}^{\star}_j
    =
    (\tilde{\mathbf{z}}_j-\boldsymbol{\mu}_z)
    \oslash
    \boldsymbol{\sigma}_z
    \label{eq:standardized_structure}
\end{equation}
be the standardized structural interaction vector. We project it into the response-coordinate space by
\begin{equation}
    \mathbf{h}_j
    =
    W\mathbf{z}^{\star}_j+\boldsymbol{\xi},
    \qquad
    \mathbf{h}_j=(h_{j,0},h_{j,v},h_{j,t},h_{j,vt})^{\top}.
    \label{eq:bridge}
\end{equation}
The projected vector \(\mathbf{h}_j\) serves as a structural anchor for the item parameters of the modality-conditioned IRT model.

Specifically, the discrimination and difficulty parameters are parameterized as
\begin{equation}
    a_{j,k}
    =
    \operatorname{softplus}
    \left(
    \alpha_k h_{j,k}+\beta_k+\Delta^{a}_{j,k}
    \right)
    +\epsilon_0,
    \quad
    b_{j,k}
    =
    \operatorname{softplus}
    \left(
    \gamma_k h_{j,k}+\eta_k+\Delta^{b}_{j,k}
    \right)
    +\epsilon_0,
    \label{eq:anchored_item_params}
\end{equation}
for \(k\in\{0,v,t,vt\}\).
Here \(\alpha_k,\beta_k,\gamma_k,\eta_k\) are global calibration parameters, while \(\Delta^{a}_{j,k}\) and \(\Delta^{b}_{j,k}\) are item-specific residuals learned from extrinsic model responses.

This parameterization implements the anchoring step in structure-response co-calibration. 
The anchor term injects intrinsic structural evidence into the response model, while the residual terms allow response-side calibration to revise the structural prior when observed model behavior disagrees with the structural estimate.

\subsection{Response-Calibrated Modality-Dependence Profile}
\label{sec:response_calibrated_profile}

After fitting the structure-anchored response model, \ourq derives a calibrated modality-dependence profile from the learned response surface. 
Let \(\bar{\boldsymbol{\theta}}\) denote the average ability vector of the diagnostic model pool. 
For each item \(j\), we compute calibrated correctness probabilities
\begin{equation}
    p_{j,s}
    =
    P(r=1\mid j,c_s,\bar{\boldsymbol{\theta}}),
    \qquad
    s\in\{0,v,t,vt\}.
    \label{eq:calibrated_probability}
\end{equation}

For notational compactness, we write \(p_s=p_{j,s}\) for a fixed item \(j\). 
We define the response-side dependence scores as
\begin{equation}
\boldsymbol{\chi}_j
=
\bigl(
\min(p_v,p_t)-p_0,\;
p_v-\max(p_t,p_0),\;
p_t-\max(p_v,p_0),\;
p_{vt}-\max(p_v,p_t)
\bigr).
\label{eq:response_scores}
\end{equation}
The four coordinates correspond to redundant, visual-dominant, textual-dominant, and synergistic dependence, respectively. 
They measure shared unimodal gain over the no-modality condition, asymmetric unimodal advantages, and the extra gain of the full multimodal condition over the best unimodal condition.

The final response-calibrated modality-dependence profile is
\begin{equation}
    \mathbf{q}^{\mathrm{cal}}_j
    =
    \operatorname{softmax}(\boldsymbol{\chi}_j).
\end{equation}
Compared with raw accuracy differences across modality conditions, \(\mathbf{q}^{\mathrm{cal}}_j\) is calibrated through latent model ability and structure-anchored item parameters, reducing sensitivity to the particular composition of the diagnostic model pool. We therefore interpret \(\mathbf{q}^{\mathrm{cal}}_j\) not only as a modality-dependence profile, but also as an attribution over evidence factors that explains which source of evidence accounts for the item's observed evaluation behavior.

\subsection{Structure-Response Co-Calibration Objective}
\label{sec:learning_objective}

\ourq is trained by combining response-side likelihood fitting with structural-prior alignment. 
The final objective is
\begin{equation}
\mathcal{L}
=
-\sum_{(i,j,c)\in\Omega}
\left[
r_{ijc}\log P_{ijc}
+
(1-r_{ijc})\log(1-P_{ijc})
\right]
+
\frac{\lambda_{\mathrm{str}}}{J}
\sum_{j=1}^{J}
w_j
\operatorname{KL}
\left(
\mathbf{q}^{\mathrm{str}}_j
\,\Vert\,
\mathbf{q}^{\mathrm{cal}}_j
\right).
\label{eq:objective}
\end{equation}
The first term calibrates the response model using observed model behavior, while the second term preserves intrinsic item structure according to the confidence weight \(w_j\). 
Thus, confident structural estimates act as stronger anchors, whereas ambiguous estimates are mainly resolved by extrinsic model responses.

After training, \ourq uses \(\mathbf{q}^{\mathrm{cal}}_j\) as the final item-level diagnostic representation. 
Its coordinates quantify whether the calibrated evidence for item \(j\) is primarily redundant, visual-dominant, textual-dominant, or synergistic. 
This yields a modality-dependence profile that preserves intrinsic item structure while being corrected by empirical response behavior.

\section{Experiments}
\label{sec:exp}

\subsection{Experimental Setup}
\label{sec:exp_setup}

\myparagraph{Benchmarks}
We evaluate on three widely used vision-language benchmarks: ScienceQA~\citep{ScienceQA}, MMMU~\citep{MMMU}, and SEED-Bench~\citep{SeedBench}. 
They cover science-oriented question answering, college-level multimodal reasoning, and general multimodal understanding, respectively.

\myparagraph{Evaluation settings}
We validate benchmark diagnosis methods under two complementary protocols. 
First, \emph{benchmark-refinement oriented validation} follows the contaminated candidate-pool setting of prior work~\citep{M3IRT}: each pool contains both clean benchmark items and synthetic low-quality items generated by swapping image or text components across questions. 
A method selects a compact subset from this pool, and the subset is evaluated by how well it preserves the model ranking induced by the full clean benchmark.

Second, \emph{human-labeled item validation} evaluates whether a method recovers the dominant evidence factor of each item. 
We construct a human-labeled diagnostic validation set from ScienceQA, MMMU, and SEED-Bench. 
Candidate items are first filtered under the original full multimodal condition to ensure human solvability, and each retained item is then annotated under controlled evidence conditions. 
This yields a human evidence-sufficiency reference in a four-factor diagnostic space: Redundant, Visual-dominant, Textual-dominant, or Synergistic. 
The validation set contains 1,000 items per benchmark and is used only for evaluation. 
For quantitative evaluation, we compare each method's dominant diagnostic factor with the human reference and report accuracy and macro-F1. 
Full annotation details are provided in Appendix~\ref{app:human_annotation}.

\myparagraph{Model pool and response collection}
For methods using extrinsic responses, we collect modality-conditioned responses from a heterogeneous pool of VLMs, including proprietary and open-weight models from GPT-series~\citep{gpt4}, Qwen-VL~\citep{qwen2.5-vl}, Gemma~\citep{gemma3}, Ministral~\citep{ministral3}, and other model families. 
The full model list, prompts, and implementation details are provided in Appendix~\ref{app:experimental_settings}.

\myparagraph{Compared methods}
We compare \ourq with six baselines in total. 
Random selects a diagnostic factor uniformly at random. 
Raw directly maps modality-conditioned correctness gaps to the four diagnostic factors. 
LSMI~\citep{LSMI} is a structure-only baseline that infers the dominant evidence factor from sample-wise multimodal interaction signals.  
IRT~\citep{irt} is a standard psychometric baseline based on item difficulty, while MIRT~\citep{mirt} extends IRT with modality-conditioned response factors. 
M3IRT~\citep{M3IRT} is a stronger multimodal IRT baseline that models modality-specific item behavior
from extrinsic responses. 
In contrast, \ourq incorporates intrinsic structural signals as soft priors and infers the dominant evidence factor from its learned response-calibrated profile.

\subsection{Main Results}

\myparagraph{Benchmark-refinement oriented validation}
We first conduct benchmark-refinement oriented validation to assess whether diagnostic scores can support reliable benchmark-based model comparison. 
In this setting, no item-level diagnostic labels are used. 
Instead, following the contaminated candidate-pool protocol of prior work~\citep{M3IRT}, we evaluate whether a diagnosis method can select compact subsets that preserve the model-ranking conclusions of the full clean benchmark.

For each benchmark, we first compute a reference model ranking using the full clean set. 
We then construct a contaminated candidate pool by combining the clean items with an equal number of synthetic low-quality items. 
Following prior work~\citep{M3IRT}, the low-quality items are generated by swapping image or text components across different questions, producing shortcut-prone or unsolvable examples while preserving the benchmark format. 
The resulting candidate pool contains 50\% clean items and 50\% synthetic low-quality items.

We compare Random selection, IRT, M3IRT, and \ourq. 
For each method, we select subsets with sizes corresponding to \(5\%\), \(10\%\), \(15\%\), \(20\%\), and \(25\%\) of the clean set size. 
The selected subset is then used to compute model accuracies and induce a subset-based model ranking. 
We measure ranking preservation by the Spearman rank correlation between this subset-induced ranking and the reference ranking obtained from the full clean benchmark:
\begin{equation}
    \rho
    =
    \operatorname{Spearman}
    \left(
    \operatorname{rank}_{\mathrm{clean}},
    \operatorname{rank}_{\mathrm{subset}}
    \right).
    \label{eq:subset_spearman}
\end{equation}
A higher \(\rho\) indicates that the selected subset better preserves the evaluation conclusion of the full clean benchmark. 
For each subset size, we repeat item-pool resampling 20 times and report the mean and standard deviation of Spearman rank correlation.

\begin{figure}[!t]
    \centering
    \includegraphics[width=\linewidth]{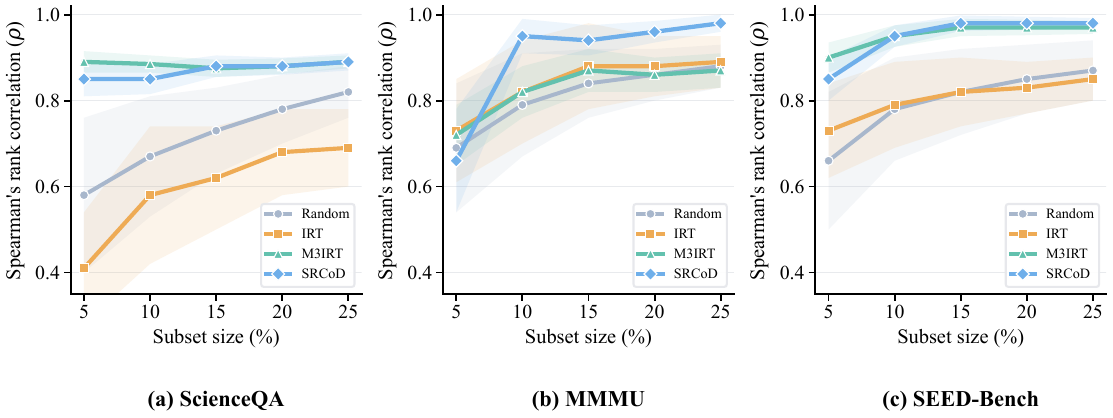}
    \caption{
    Benchmark-refinement oriented validation under contaminated candidate pools. 
    Each method selects compact subsets from candidate pools containing equal numbers of clean benchmark items and synthetic low-quality items. 
    We report the Spearman rank correlation between the model ranking induced by the selected subset and the ranking induced by the full clean benchmark. 
    }
    \label{fig:robust_subset_selection}
\end{figure}

Figure~\ref{fig:robust_subset_selection} shows that \ourq achieves strong benchmark-refinement oriented validation performance under contaminated candidate pools. 
On ScienceQA, \ourq maintains high rank correlation across all subset sizes and is consistently stronger than IRT and Random, while remaining competitive with M3IRT. 
On MMMU, \ourq achieves the best performance for subset sizes of \(10\%\) and above, with rank correlation approaching the full clean benchmark ranking even with relatively small subsets. 
On SEED-Bench, \ourq and M3IRT both perform strongly, but \ourq reaches near-ceiling rank correlation from \(10\%\) onward and remains substantially better than IRT and Random. 
Overall, these results indicate that the diagnostic profiles produced by \ourq can identify items that preserve reliable model-comparison conclusions even when the candidate pool is contaminated.

\myparagraph{Human-labeled item validation}
We further evaluate whether the diagnostic profile identifies the correct dominant evidence factor using the human-labeled validation set. 
The comparison among baselines highlights the complementary roles of intrinsic item structure and extrinsic model responses. 
The structure-only LSMI baseline performs above random assignment in accuracy but obtains relatively low macro-F1, indicating that intrinsic structural signals capture some dominant dependence patterns but are insufficient for balanced diagnosis across all categories. 
Extrinsic-response baselines generally perform better, showing that modality-conditioned responses provide useful empirical evidence for inferring the dominant evidence factor.
However, their performance still lags behind \ourq, suggesting that response evidence alone does not fully resolve fine-grained modality-dependence diagnosis.

\ourq consistently outperforms the baselines by combining the two sources of evidence. 
The structural prior preserves intrinsic modality-dependence information, while response-side calibration adjusts this prior using observed modality-conditioned responses. 
These results support the central motivation of \ourq: reliable multimodal benchmark diagnosis benefits from structure-response co-calibration rather than relying solely on intrinsic item structure or extrinsic model responses.

\begin{table*}[t]
    \centering
    \caption{
    Human-labeled item validation.
    Accuracy and macro-F1 measure agreement with human evidence-sufficiency references in the four-factor diagnostic space.
    Bold and underline denote the best and second-best results, respectively.
    }
    \vspace{1.0em}
    \label{tab:main_results}
    \resizebox{0.9\textwidth}{!}{
    \begin{tabular}{lcccccccc}
        \toprule
        \multirow{2}{*}{Method}
        & \multicolumn{2}{c}{ScienceQA}
        & \multicolumn{2}{c}{MMMU}
        & \multicolumn{2}{c}{SEED-Bench}
        & \multicolumn{2}{c}{Average} \\
        \cmidrule(lr){2-3}
        \cmidrule(lr){4-5}
        \cmidrule(lr){6-7}
        \cmidrule(lr){8-9}
        & Acc. & Macro-F1
        & Acc. & Macro-F1
        & Acc. & Macro-F1
        & Acc. & Macro-F1 \\
        \midrule
        Random
        & 0.251 & 0.233
        & 0.249 & 0.241
        & 0.250 & 0.242
        & 0.250 & 0.239 \\

        Raw
        & 0.353 & 0.334
        & \underline{0.503} & 0.435	
        & 0.388 & 0.355	
        & 0.415 & 0.375 \\

        LSMI
        & 0.464 & 0.255
        & 0.265 & 0.171
        & 0.309 & 0.189
        & 0.346 & 0.205 \\

        IRT
        & 0.507 & 0.476
        & 0.384 & 0.382
        & 0.494 & 0.474
        & 0.462 & 0.444 \\

        MIRT
        & 0.506 & 0.476
        & 0.371 & 0.370
        & 0.494 & 0.473
        & 0.457 & 0.440 \\

        M3IRT
        & \underline{0.570} & \underline{0.538}
        & 0.486 & \underline{0.484}
        & \underline{0.618} & \underline{0.603}
        & \underline{0.558} & \underline{0.542} \\

        \rowcolor{blue!10}
        \ourq
        & \textbf{0.607} & \textbf{0.588}
        & \textbf{0.548} & \textbf{0.538}
        & \textbf{0.685} & \textbf{0.673}
        & \textbf{0.613} & \textbf{0.600} \\
        \bottomrule
    \end{tabular}
    }
\end{table*}

\begin{figure}[t]
    \centering
    \includegraphics[width=\linewidth]{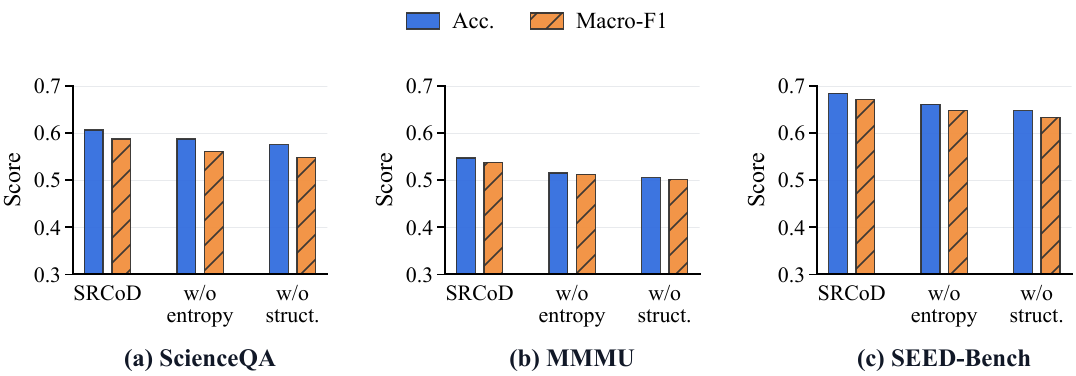}
    \caption{
    Ablation study of \ourq. We compare the full model with variants that remove
    entropy-based confidence weighting or the profile-level structural alignment
    loss. 
    }
    \label{fig:ablation}
\end{figure}

\subsection{Ablation Study}
\label{sec:ablation}

We conduct an ablation study to examine two key components of \ourq: entropy-based confidence weighting and profile-level structural alignment. The \textit{w/o entropy} variant removes the entropy-based weight \(w_j\), treating all structural priors as equally reliable. The \textit{w/o struct.} variant removes the structural alignment loss \(\mathcal{L}_{\mathrm{str}}\), while keeping the response-side model and structure-anchored item parameterization unchanged.

Figure~\ref{fig:ablation} shows that the full \ourq model consistently achieves the best accuracy and macro-F1 across the three benchmarks. Removing entropy weighting degrades performance, indicating that uncertain structural priors should not be enforced as strongly as confident ones. Entropy weighting therefore improves robustness by down-weighting ambiguous intrinsic structural signals. Removing profile-level structural alignment also reduces performance. This suggests that intrinsic structural signals are most effective when they both anchor item parameters and regularize the final response-calibrated diagnostic profile.

Overall, the ablation results show that the two components play complementary roles: entropy weighting controls the reliability of structural priors, while profile-level alignment transfers intrinsic structural information to the final diagnostic profile.

\begin{figure}[!t]
    \centering
    \includegraphics[width=\linewidth]{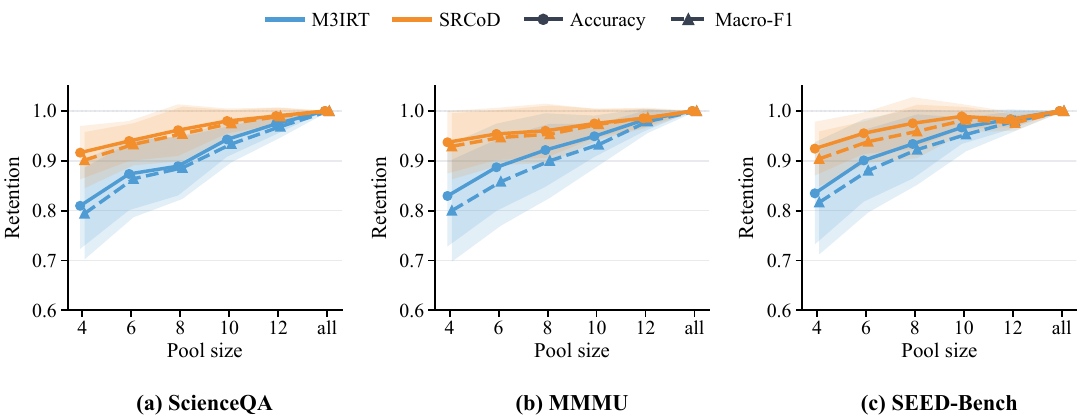}
    \caption{
    Robustness to model-pool subsampling. For each pool size, we randomly sample diagnostic models from the full model pool and repeat the experiment 10 times.
    We report performance retention relative to the full-pool setting. Shaded
    regions denote one standard deviation.
    }
    \label{fig:model_pool_subsampling}
\end{figure}

\subsection{Robustness to Model-Pool Subsampling}
\label{sec:model_pool_subsampling}

A reliable benchmark diagnosis method should produce stable item-level conclusions under different diagnostic model pools. This is important for multimodal benchmark auditing, because collecting modality-conditioned responses from a large pool of VLMs can be costly. We therefore validate robustness when only a subset of diagnostic models is available.

For each benchmark, we randomly sample \(n \in \{4,6,8,10,12\}\) models from the full diagnostic model pool. For each sampled pool, we fit the diagnosis method using only the selected models and validate item-level predictions on the same human-labeled validation set. We repeat this procedure 10 times for each pool size and report performance retention relative to the full-pool setting:
\begin{equation}
    \mathrm{Retention}^{(s)}(n)
    =
    \frac{
    \mathrm{Score}^{(s)}(n)
    }{
    \mathrm{Score}(\mathrm{all})
    },
    \label{eq:model_pool_retention}
\end{equation}
where \(s\) indexes the subsampling trial, and \(\mathrm{Score}\) denotes either accuracy or macro-F1. Figure~\ref{fig:model_pool_subsampling} reports the mean retention over 10 trials, with shaded regions denoting one standard deviation.

\begin{wrapfigure}{r}{0.5\linewidth}
    \vspace{-1.2em}
    \centering
    \includegraphics[width=\linewidth]{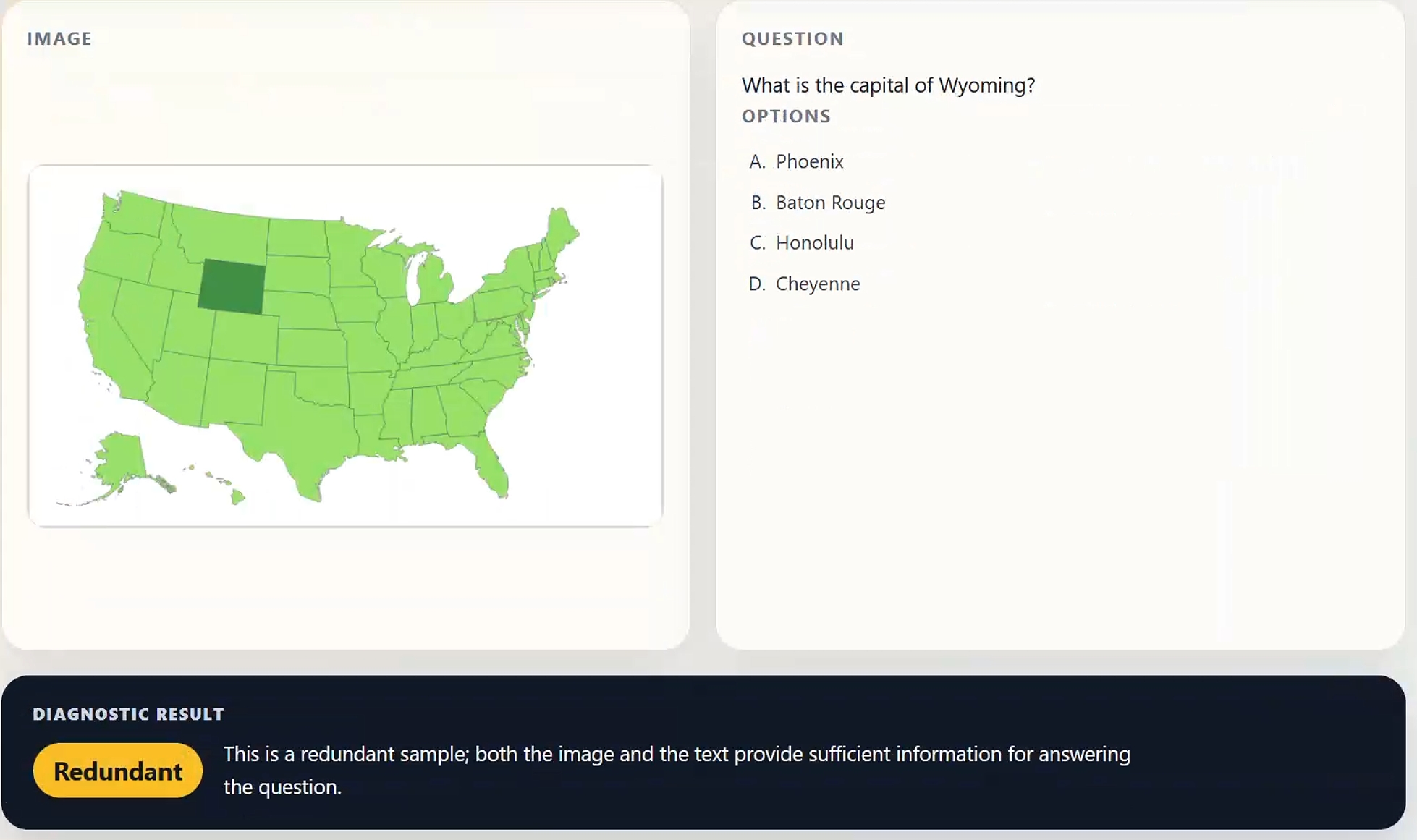}
    \caption{
    Qualitative example of the diagnostic output produced by \ourq. 
    The item is diagnosed as \textit{Redundant}, indicating that both the image and the text provide sufficient evidence for answering the question.
    }
    \label{fig:qualitative_diagnostic}
\end{wrapfigure}

Figure~\ref{fig:model_pool_subsampling} shows that \ourq is more robust than M3IRT under small diagnostic model pools. When only four or six models are used, M3IRT shows a clear drop in accuracy and macro-F1 retention, whereas \ourq stays closer to its full-pool performance across the three benchmarks. \ourq also exhibits smaller variation across subsampling trials, indicating that intrinsic structural anchoring reduces sensitivity to the particular choice of diagnostic models.

Overall, these results provide additional evidence for structure-anchored diagnosis. By combining extrinsic model responses with intrinsic item structure, \ourq produces more stable modality-dependence estimates when the diagnostic model pool is limited.

We further validate the behavioral generalization of \ourq on held-out model families in Appendix~\ref{app:heldout_families}, where the predicted diagnostic categories consistently exhibit their expected modality-dependent response patterns.

\subsection{Qualitative Diagnostic Example}

Figure~\ref{fig:qualitative_diagnostic} presents a representative example of the diagnostic output produced by \ourq.
The question asks for the capital of Wyoming, while the accompanying map highlights Wyoming in dark green.
The correct answer is \textit{Cheyenne}.
From the textual input alone, a model can identify the state directly from the question and retrieve its capital using geographic knowledge, without consulting the image.
Meanwhile, the visual input provides an alternative way to identify the state through the highlighted region on the map.
Consequently, both modalities convey sufficient information to establish the geographic entity needed to answer the question.
Based on this overlapping evidence, \ourq classifies the item as \textit{Redundant}.
This example demonstrates that \ourq provides an interpretable diagnostic result by identifying how the available evidence contributes to answerability.

\section{Conclusion}
This paper studied item-level multimodal benchmark diagnosis. We proposed \ourq, a structure-anchored multimodal IRT framework that co-calibrates intrinsic item structure with extrinsic model responses. By using sample-wise multimodal interaction signals as structural priors and calibrating them with modality-conditioned response behavior, \ourq produces fine-grained diagnostic profiles for redundant, visual-dominant, textual-dominant, and synergistic items. We also introduced a controlled human validation protocol and constructed a diagnostic validation set across ScienceQA, MMMU, and SEED-Bench. Experiments show that \ourq outperforms intrinsic-structure and extrinsic-response baselines in item-level diagnosis, remains more stable under limited model pools, and preserves model rankings when selecting compact benchmark subsets. Future work may extend this framework to open-ended tasks and broader multimodal settings beyond vision and language.

\begin{ack}
This work was partially supported by the National Natural Science Foundation of China (Grant Nos. U23A20296, 62272469 and 62302513), and the Science and Technology Innovation Program of Hunan Province (2026RC3144).
\end{ack}

\bibliographystyle{unsrtnat}
\balance
\bibliography{simple-base}


\appendix

\section{Limitation}

This work has three main limitations.
First, our experiments focus on multiple-choice vision-language benchmarks. Although ScienceQA, MMMU, and SEED-Bench cover diverse domains and reasoning types, \ourq has not yet been validated on open-ended generation tasks, interactive multimodal evaluation, or modalities beyond vision and language.

Second, \ourq relies on a diagnostic pool of VLMs to collect modality-conditioned responses. While structural anchoring improves robustness, the resulting profiles remain sensitive to the composition and capabilities of the selected models. For example, limited understanding of molecular diagrams may lead to overestimated cross-modal synergy, while difficulties in interpreting contingency tables may cause visual contributions to be underestimated.

Third, both the intrinsic structural prior and the human validation labels involve uncertainty. Sample-wise multimodal interaction estimates are approximations of item-level evidence structure and may be affected by representation quality. Meanwhile, controlled single-modality human labels can depend on annotator background knowledge and item ambiguity. 

These limitations suggest that \ourq provides estimates of modality dependence rather than definitive judgments of human evidence sufficiency. Future work may explore broader evaluation settings, more representative diagnostic model pools, and graded evidence-sufficiency annotations.

\section{Preliminaries}
\label{sec:background_related_work}

\subsection{Partial Information Decomposition}
\label{sec:pid}

Partial Information Decomposition (PID) provides a framework for analyzing how multiple information sources contribute to a target variable. Given two sources \(X^{v}\) and \(X^{t}\), and a target variable \(Y\), PID decomposes the target-relevant information into four components:
\begin{equation}
    (R, U_v, U_t, S),
\end{equation}
where \(R\) denotes redundant information shared by both sources, \(U_v\) and \(U_t\) denote source-specific information uniquely provided by \(X^{v}\) and \(X^{t}\), respectively, and \(S\) denotes synergistic information that is available only from the joint use of both sources.

In the two-source case, these components satisfy
\begin{equation}
    I(X^{v};Y)=R+U_v,
    \qquad
    I(X^{t};Y)=R+U_t,
    \qquad
    I(X^{v},X^{t};Y)=R+U_v+U_t+S.
    \label{eq:pid_decomposition}
\end{equation}

In vision-language evaluation, the two sources naturally correspond to the visual input and the textual input, while the target corresponds to the answer. This makes PID a useful conceptual tool for distinguishing whether answer-relevant information is available redundantly from either modality, uniquely from one modality, or only through their interaction.

\subsection{Item Response Theory and Multimodal Extensions}
\label{sec:mirt}

Item Response Theory (IRT) models binary responses as a function of subject ability and item properties. In benchmark evaluation, models can be treated as subjects and benchmark examples as items. The standard two-parameter logistic model is
\begin{equation}
    P(r_{ij}=1)
    =
    \sigma\left(a_j(\theta_i-b_j)\right),
    \label{eq:standard_irt}
\end{equation}
where \(\theta_i\) denotes the ability of subject \(i\), and \(a_j\) and \(b_j\) denote the discrimination and difficulty of item \(j\).

Multidimensional IRT (MIRT) generalizes IRT by representing ability and discrimination in a latent vector space:
\begin{equation}
    P(r_{ij}=1)
    =
    \sigma\left(\mathbf{a}_j^\top \boldsymbol{\theta}_i-b_j\right),
    \label{eq:mirt}
\end{equation}
where \(\boldsymbol{\theta}_i\) is a multidimensional ability vector and \(\mathbf{a}_j\) is a multidimensional discrimination vector.

For vision-language evaluation, multimodal IRT further conditions the response process on modality availability. It decomposes model ability and item properties into modality-related coordinates, such as base, visual, textual, and joint visual-textual components. \ourq adopts this modality-conditioned response modeling principle, but differs from response-only approaches by anchoring item parameters with intrinsic item structure.

\section{Human Annotation Protocol}
\label{app:human_annotation}

This section describes the construction of the human-labeled diagnostic validation set used in Section~\ref{sec:exp}. 
The validation set is designed to provide an independent evaluation signal for item-level multimodal benchmark diagnosis. 
It is used only for evaluation and is not used to train \ourq, tune its hyperparameters, or fit any baseline method.

\subsection{Candidate Items}
\label{app:candidate_items}

We construct the validation set from three benchmarks: ScienceQA~\citep{ScienceQA}, MMMU~\citep{MMMU}, and SEED-Bench~\citep{SeedBench}. 
Candidate items are sampled from their multiple-choice vision-language question answering subsets. 
Each item consists of a visual input, a textual question, a set of answer options, and the ground-truth answer.

To ensure that diagnostic labels reflect modality dependence rather than item invalidity, we first filter candidate items under the original full multimodal condition. 
In this condition, annotators are given the original image, question, and answer options. 
Each candidate item is answered by \(M=3\) annotators. 
Let \(r^{f}_{mj}\in\{0,1\}\) denote whether annotator \(m\) answers item \(j\) correctly under the full multimodal condition. 
We retain item \(j\) only if a majority of annotators answer it correctly:
\begin{equation}
    s_f(j)
    =
    \mathbb{I}
    \left[
    \sum_{m=1}^{M} r^{f}_{mj}
    \geq
    \left\lceil \frac{M}{2} \right\rceil
    \right].
    \label{eq:app_full_filter}
\end{equation}
Items with \(s_f(j)=1\) form the retained pool. 
From this pool, we sample 1,000 items per benchmark to form the diagnostic validation set.

\subsection{Controlled Evidence Conditions}
\label{app:controlled_conditions}

For each retained item, we collect human answers under two controlled evidence conditions. 
These conditions are designed to test whether a single modality provides sufficient evidence for solving the item. 
To reduce carryover effects, an annotator is assigned at most one controlled evidence condition for a given item, and item order is randomized within each annotation batch. 
Annotators are instructed to select exactly one option for each item and are not allowed to access external resources.

\myparagraph{Visual-conditioned setting}
Annotators are provided with the visual input and answer options. 
When necessary, we provide a minimal task instruction that preserves the task format while removing answer-bearing textual evidence. 
For example, the instruction may specify that the annotator should select the option best supported by the image, but it does not include object names, relations, quantities, or other cues that would reveal the answer without visual evidence. 
This condition is used to test whether the item can be solved primarily from visual evidence.

\myparagraph{Textual-conditioned setting}
Annotators are provided with the textual question and answer options, but not the visual input. 
This condition is used to test whether the item can be solved primarily from textual cues or language priors.

Each retained item is answered by \(M=5\) annotators under each controlled evidence condition. 
The annotator index is condition-specific and does not imply that the same annotator sees the same item under both controlled conditions.

\subsection{Label Assignment}
\label{app:label_assignment}

Let \(r^{v}_{mj}\in\{0,1\}\) denote whether annotator \(m\) answers item \(j\) correctly under the visual-conditioned setting, and let \(r^{t}_{mj}\in\{0,1\}\) denote whether annotator \(m\) answers item \(j\) correctly under the textual-conditioned setting. 
We define two binary sufficiency indicators:
\begin{equation}
    s_v(j)
    =
    \mathbb{I}
    \left[
    \sum_{m=1}^{M} r^{v}_{mj}
    \geq
    \left\lceil \frac{M}{2} \right\rceil
    \right],
    \qquad
    s_t(j)
    =
    \mathbb{I}
    \left[
    \sum_{m=1}^{M} r^{t}_{mj}
    \geq
    \left\lceil \frac{M}{2} \right\rceil
    \right].
    \label{eq:app_sufficiency}
\end{equation}
A controlled evidence condition is therefore considered sufficient if a majority of annotators answer the item correctly under that condition.

Each item is assigned one of four modality-dependence labels:
\begin{equation}
y_j =
\begin{cases}
\text{Redundant}, & s_v(j)=1,\ s_t(j)=1,\\
\text{Visual-dominant}, & s_v(j)=1,\ s_t(j)=0,\\
\text{Textual-dominant}, & s_v(j)=0,\ s_t(j)=1,\\
\text{Synergistic}, & s_v(j)=0,\ s_t(j)=0.
\end{cases}
\label{eq:app_human_label}
\end{equation}

The Redundant label indicates that either controlled evidence condition is sufficient for human annotators to answer the item. 
The Visual-dominant and Textual-dominant labels indicate that the item is primarily supported by visual or textual evidence, respectively. 
The Synergistic label indicates that neither controlled evidence condition is sufficient, while the item remains human-solvable under the original full multimodal condition.

\subsection{Annotation Reliability}
\label{app:annotation_reliability}

We report inter-annotator reliability using Fleiss' Kappa~\citep{kapas} to quantify the consistency of human judgments. 
For the full multimodal filtering stage, Fleiss' Kappa is 0.76, indicating substantial agreement among annotators. 
For the two controlled evidence conditions, Fleiss' Kappa is 0.56 for the visual-conditioned setting and 0.63 for the textual-conditioned setting. 
These values indicate moderate agreement, reflecting the greater ambiguity of judging item answerability from a single modality. 
Overall, the annotation results suggest that the retained full-condition items are consistently human-solvable, while the controlled single-modality labels provide reliable but non-trivial diagnostic signals.

\subsection{Label Distribution of the Diagnostic Validation Set}
\label{app:label_distribution}

\begin{figure}[t]
    \centering
    \includegraphics[width=\linewidth]{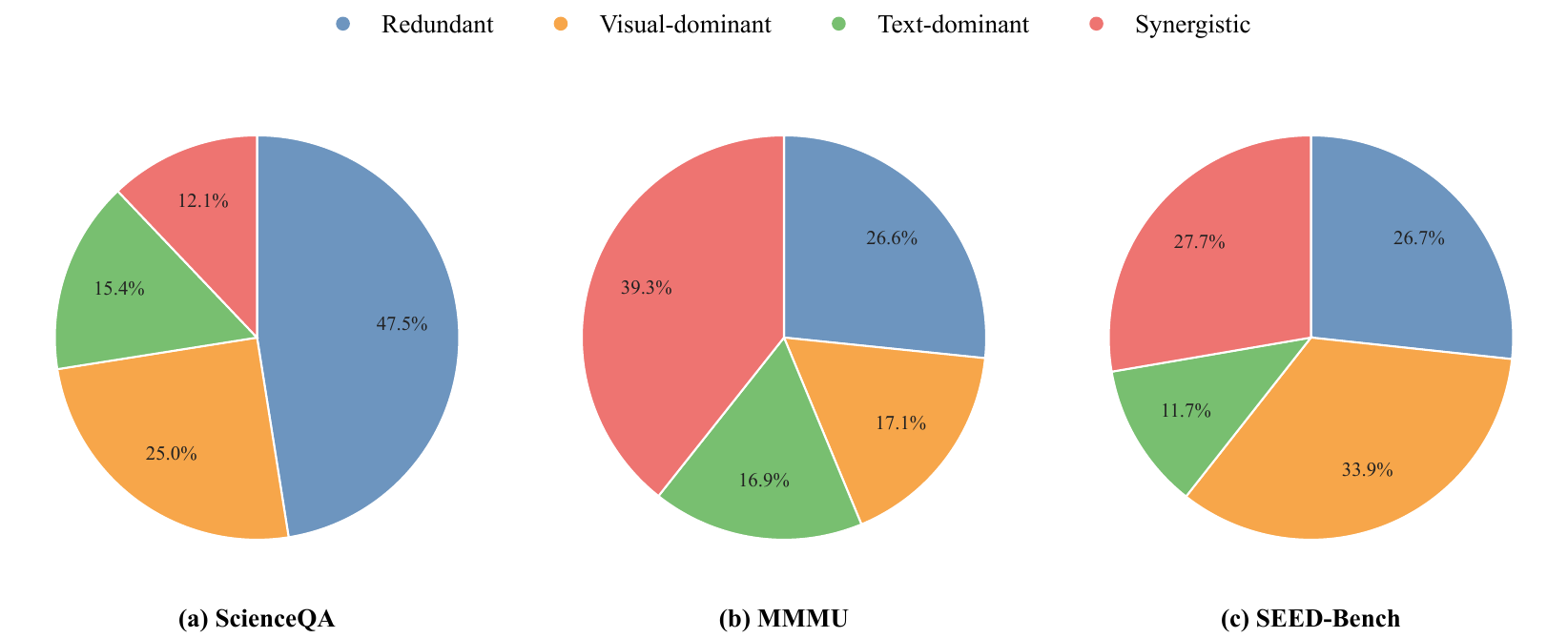}
    \caption{Distribution of human-annotated modality-dependence labels across ScienceQA, MMMU, and SEED-Bench. 
    Each pie chart shows the proportion of Redundant, Visual-dominant, Textual-dominant, and Synergistic items in the corresponding benchmark.}
    \label{fig:app_label_distribution}
\end{figure}

Figure~\ref{fig:app_label_distribution} summarizes the modality-dependence label distribution of the human-labeled diagnostic validation set. 
The proportions of Redundant, Visual-dominant, Textual-dominant, and Synergistic items vary across ScienceQA, MMMU, and SEED-Bench, indicating that different benchmarks contain different mixtures of shortcut-prone and cross-modally dependent items. 
This variation further motivates item-level benchmark diagnosis rather than assuming a fixed dataset-level modality structure.

\section{Details of Experimental Settings}
\label{app:experimental_settings}

\subsection{Evaluation Metric}
\label{app:evaluation_details}

We evaluate all methods against the human-labeled diagnostic validation set. 
Each item \(j\) has a human annotation label 
\(y_j \in \mathcal{Y}\), where
\begin{equation}
\mathcal{Y}
=
\{\text{Redundant}, \text{Visual-dominant}, \text{Textual-dominant}, \text{Synergistic}\}.
\end{equation}
For methods that output a four-dimensional modality-dependence profile, we predict the label by taking the coordinate with the largest score:
\begin{equation}
\hat{y}_j
=
\arg\max_{c\in\mathcal{Y}}
q_{j,c}.
\end{equation}
For methods that output modality-conditioned scores or probabilities, we first map their outputs to the same four-factor diagnostic space and then apply the same argmax rule.

We report accuracy and macro-F1 as the main evaluation metrics. 
Accuracy measures the overall fraction of correctly predicted labels:
\begin{equation}
    \mathrm{Acc}
    =
    \frac{1}{N}
    \sum_{j=1}^{N}
    \mathbb{I}
    \left[
    \hat{y}_j = y_j
    \right],
    \label{eq:app_accuracy}
\end{equation}
where \(N\) is the number of items in the diagnostic validation set.

To compute macro-F1, we first compute precision, recall, and F1 for each class \(c\in\mathcal{Y}\). 
Let \(\mathrm{TP}_c\), \(\mathrm{FP}_c\), and \(\mathrm{FN}_c\) denote the number of true positives, false positives, and false negatives for class \(c\). 
The class-wise precision and recall are
\begin{equation}
    \mathrm{Prec}_c
    =
    \frac{\mathrm{TP}_c}
    {\mathrm{TP}_c+\mathrm{FP}_c},
    \qquad
    \mathrm{Rec}_c
    =
    \frac{\mathrm{TP}_c}
    {\mathrm{TP}_c+\mathrm{FN}_c}.
    \label{eq:app_precision_recall}
\end{equation}
The F1 score for class \(c\) is then
\begin{equation}
    \mathrm{F1}_c
    =
    \frac{
    2\,\mathrm{Prec}_c\,\mathrm{Rec}_c
    }{
    \mathrm{Prec}_c+\mathrm{Rec}_c
    }.
    \label{eq:app_class_f1}
\end{equation}
Finally, macro-F1 is computed by averaging the class-wise F1 scores over the four diagnostic categories:
\begin{equation}
    \mathrm{Macro\text{-}F1}
    =
    \frac{1}{|\mathcal{Y}|}
    \sum_{c\in\mathcal{Y}}
    \mathrm{F1}_c.
    \label{eq:app_macro_f1}
\end{equation}

We use macro-F1 in addition to accuracy because the human-labeled diagnostic set is not perfectly class-balanced. 
Macro-F1 gives equal weight to Redundant, Visual-dominant, Textual-dominant, and Synergistic items, making it more informative for evaluating whether a method performs consistently across all modality-dependence categories.

\subsection{Computational Resources}
\label{app:computational_resources}

The computational resources used in this study are summarized in Table~\ref{tab:computational_resources}. All modality-conditioned evaluations  estimation are conducted on the same computing environment. For each benchmark, we collect model responses under the required modality conditions and then fit the corresponding diagnostic models.

\begin{table}[t]
\centering
\caption{Computational resources used for experiments.}
\label{tab:computational_resources}
\begin{tabular}{ll}
\toprule
Component & Specification \\
\midrule
CPU & 12th Gen Intel(R) Core(TM) i7-12700KF @ 3.60 GHz \\
RAM & 32 GB \\
GPU & NVIDIA GeForce RTX 4090, 24 GB \\
Python version & 3.10.1 \\
PyTorch version & 2.7.0 \\
CUDA version & 12.6 \\
\bottomrule
\end{tabular}
\end{table}

\subsection{Datasets}
\label{app:datasets}

We use three vision-language benchmarks in our experiments.

ScienceQA~\cite{ScienceQA} is a science-oriented multimodal question answering benchmark. It contains multiple-choice questions with textual contexts and, when available, visual inputs. The license for the ScienceQA dataset is Creative Commons Attribution-NonCommercial-ShareAlike 4.0 International.

MMMU~\cite{MMMU} is a massive multi-discipline multimodal understanding and reasoning benchmark designed to evaluate college-level reasoning ability across diverse domains, including science, medicine, engineering, business, humanities, and art. The license for the MMMU dataset is Apache License 2.0.

SEED-Bench~\cite{SeedBench} is a large-scale benchmark for evaluating general multimodal understanding abilities using multiple-choice questions. We use its vision-language question answering items in our experiments. The official release materials indicate that SEED-Bench is released under Apache License 2.0, while the Hugging Face dataset card lists Attribution-NonCommercial 4.0 International and notes that some underlying visual assets are subject to their original dataset licenses.

\subsection{Models}
\label{app:models}

We use a diverse pool of multimodal large language models as diagnostic respondents. 
As summarized in Table~\ref{tab:ai-models-used}, the pool includes both closed-source API models and open-weight models, covering different model families, parameter scales, and deployment settings. 
Closed models provide strong contemporary VLM baselines, while open-weight models increase transparency and make it possible to examine performance variation across model sizes. 
This heterogeneous model pool is used to collect modality-conditioned responses for fitting the response-side components of \ourq and the compared response-based baselines. 
No model in the pool is fine-tuned during our experiments; all models are evaluated under the same benchmark prompts and modality conditions.

\begin{table}[ht]
\centering
\caption{Overview of AI Models Used}
\label{tab:ai-models-used}
\begin{tabular}{lll}
\toprule
\textbf{Model Name} & \textbf{Type} & \textbf{License or Terms} \\
\midrule
\texttt{gpt-4.1}~\citep{gpt4} & Closed & OpenAI Terms of Use \\
\texttt{gpt-4.1-nano}~\citep{gpt4} & Closed & OpenAI Terms of Use \\
\texttt{gpt-4o-mini}~\citep{gpt4} & Closed & OpenAI Terms of Use \\
\texttt{gemini-3.1-pro-preview}~\citep{gemini} & Closed & Gemini API Additional Terms of Service \\
\texttt{gemini-3.1-flash-lite}~\citep{gemini} & Closed & Gemini API Additional Terms of Service \\
\texttt{gemini-2.5-flash-lite}~\citep{gemini} & Closed & Gemini API Additional Terms of Service 
\\
\texttt{llava:13b}~\citep{llava} & Open-weight & Llama 2 Community License Agreement \\
\texttt{llava:7b}~\citep{llava} & Open-weight & Apache 2.0 \\
\texttt{gemma3:12b}~\citep{gemma3} & Open-weight & Gemma Terms of Use \\
\texttt{gemma3:4b}~\citep{gemma3} & Open-weight & Gemma Terms of Use \\
\texttt{ministral-3:14b}~\citep{ministral3} & Open-weight & Apache 2.0 \\
\texttt{ministral-3:8b}~\citep{ministral3} & Open-weight & Apache 2.0 \\
\texttt{qwen2.5vl:7b}~\citep{qwen2.5-vl} & Open-weight & Apache 2.0 \\
\texttt{qwen2.5vl:3b}~\citep{qwen2.5-vl}& Open-weight & Apache 2.0  \\
\bottomrule
\end{tabular}
\end{table}

\subsection{Prompts}
\label{app:prompts}

We use standardized multiple-choice prompts to collect model responses under different modality conditions. 
For all models and benchmarks, the prompt asks the model to select one answer option and to output only the option letter. 
We do not ask models to provide explanations or chain-of-thought rationales, since the response-side model only requires binary correctness labels. 
For each item, we evaluate the model under the modality conditions used in Section~\ref{sec:method}, while keeping the answer options and output format fixed across conditions.

Table~\ref{tab:app_prompt_conditions} summarizes the information exposed to the model under each condition.

\begin{table}[t]
\centering
\caption{Input information provided under each modality condition.}
\label{tab:app_prompt_conditions}
\begin{tabular}{lccc}
\toprule
Condition & Image & Question text & Answer options \\
\midrule
Full multimodal & Yes & Yes & Yes \\
Visual-conditioned & Yes & No & Yes \\
Textual-conditioned & No & Yes & Yes \\
No-modality & No & No & Yes \\
\bottomrule
\end{tabular}
\end{table}

The full multimodal prompt is used to measure model performance under the original benchmark setting:
\begin{quote}
\small

You are solving a multiple-choice question. Output exactly one uppercase letter from [A, B, C, ...] and nothing else.

Mode: image + text.

Use both the provided image(s) and question text.

Question: \{question\}

Options: \{options\}

Answer:

\end{quote}

The visual-conditioned prompt removes the original question text while retaining the image and answer options:
\begin{quote}
\small
You are solving a multiple-choice question. Output exactly one uppercase letter from [A, B, C, ...] and nothing else.

Mode: image only.

Question text is hidden. Use only the provided image(s) and options.

Options: \{options\}

Answer:
\end{quote}

The textual-conditioned prompt removes the image while retaining the original question and answer options:
\begin{quote}
\small
You are solving a multiple-choice question. Output exactly one uppercase letter from [A, B, C, ...] and nothing else.

Mode: text only.

No image is available. Use only question text and options.

Question: \{question\}

Options: \{options\}

Answer:
\end{quote}

The no-modality prompt removes both the image and the question text while retaining the answer options:
\begin{quote}
\small
You are solving a multiple-choice question. Output exactly one uppercase letter from [A, B, C, ...] and nothing else.

Mode: none.

No image and no question text are available.

Choose the most plausible answer only from the options below. Guess if needed.

Options: \{options\}

Answer:
\end{quote}

For all prompts, \{question\} denotes the original benchmark question and \{options\} denotes the answer options in their original order. 
We use deterministic decoding when supported by the model API or inference backend. 
After generation, we extract the first valid option letter from the model output. 
If no valid option letter can be extracted, the response is treated as incorrect. 
The extracted answer is compared against the ground-truth answer to obtain the binary response used by the response model.

\subsection{Implementation details}
We implement \ourq using Pytorch~\citep{Pytorch}. For the structural side, we use the LSMI coordinates \((R,U_v,U_t,S)\). Each coordinate is first clipped at the 0.5-th and 99.5-th percentiles and then standardized across items before being projected into the response-coordinate space. The LSMI-side profile prior is constructed from robustly normalized answerability scores and is used together with the entropy-based confidence weight in the profile-level structural alignment term.

For the response side, we use four latent coordinates corresponding to the base, visual, textual, and joint visual-textual components. In the bounded parameterization, model abilities are constrained to be positive, with the base ability restricted to \([10^{-4},0.1]\) and the remaining modality-related abilities restricted to \([10^{-4},4]\). Item discriminations are obtained by a softplus transformation and clipped to \([10^{-4},4]\). This parameterization keeps the latent response surface numerically stable while preserving the intended interpretation of modality-specific gains.

We optimize the resulting objective with Adam using learning rate \(0.01\), batch size \(2048\), and \(300\) epochs. All experiments use a fixed random seed of \(42\).

\section{Case Study}
\subsection{Redundant Items}
\label{app:case_redundant}

\begin{figure}[t]
    \centering
    \includegraphics[width=\linewidth]{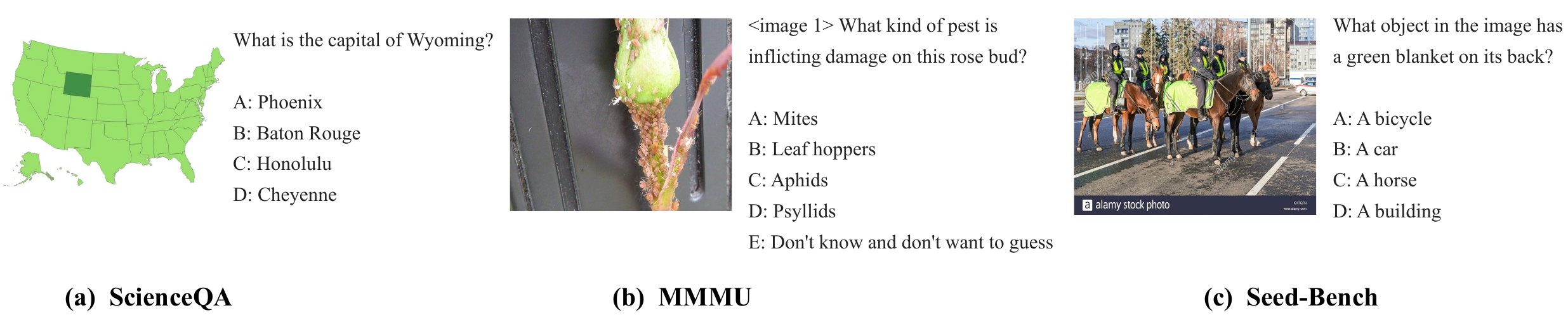}
    \caption{
    Examples of redundant items from ScienceQA, MMMU, and SEED-Bench. 
    }
    \label{fig:case_redundant}
\end{figure}

Figure~\ref{fig:case_redundant} presents examples labeled as \textit{Redundant}. 
In our annotation protocol, a redundant item is one for which both controlled views are sufficient for human annotators to select the correct answer. 
The visual-conditioned view contains the image and answer options, while the textual-conditioned view contains the original question and answer options. 
Thus, redundancy should be interpreted as controlled evidence sufficiency rather than semantic equivalence between modalities: the two views may support the same answer through different cues.

\myparagraph{ScienceQA}
The ScienceQA example asks for the capital of Wyoming, with options Phoenix, Baton Rouge, Honolulu, and Cheyenne. 
The textual-conditioned view is sufficient because the question explicitly specifies Wyoming, and the options are candidate city names. 
A human annotator with the relevant factual knowledge can identify Cheyenne as the capital.

The visual-conditioned view is also sufficient because the map highlights Wyoming. 
Although the original question is not shown in this view, the highlighted state and the city-name options strongly indicate that the item concerns the state shown on the map. 
Since the options are U.S. state capitals, the highlighted state can be matched to Cheyenne. 
Therefore, both controlled views provide enough evidence to recover the answer, and the item does not require joint reasoning over image and text.

\myparagraph{MMMU}
The MMMU example asks what kind of pest is damaging a rose bud, with options including mites, leafhoppers, aphids, and psyllids. 
The visual-conditioned view provides a direct visual cue: the image shows a rose bud covered with many small clustered insects, a pattern characteristic of aphid infestation. 
Together with the pest-category options, this is sufficient for selecting aphids.

The textual-conditioned view supports the answer in a different way. 
The question specifies that the damaged object is a rose bud and restricts the candidates to common plant pests. 
Among the listed choices, aphids are a highly salient pest associated with rose buds. 
Thus, the answer can be inferred from the question, options, and domain priors even without relying on the image.

This example shows why redundant items are not necessarily cases where the two modalities contain the same information. 
Here, the visual view supports the answer through visible pest appearance, while the textual view supports it through semantic constraints and prior knowledge. 
Both views are sufficient, but for different reasons.

\myparagraph{SEED-Bench}
The SEED-Bench example asks which object in the image has a green blanket on its back, with options bicycle, car, horse, and building. 
The textual-conditioned view is sufficient because the phrase ``has a green blanket on its back'' strongly constrains the plausible answer: among the options, horse is the only object that naturally has a back and can plausibly carry a blanket.

The visual-conditioned view is sufficient for a different reason. 
In this view, the annotator does not see the original question, and therefore cannot know that the intended query concerns a green blanket. 
Nevertheless, the image is dominated by mounted horses, which are the most salient objects among the answer options. 
Although other option categories such as buildings may also appear in the background, they are not the central objects depicted in the image. 
Thus, the image-option alignment strongly favors horse even without the original question text.

\myparagraph{Summary}
This example illustrates a shortcut-like form of redundancy. 
The item is labeled redundant because both controlled views allow annotators to recover the correct option, but the two views do so through different mechanisms: the textual-conditioned view uses the semantic constraint introduced by the question, whereas the visual-conditioned view relies on salient object-option alignment. 
Therefore, the item can be answered without genuine cross-modal integration, even though the original full item appears to require matching a textual attribute to a visual object.

\subsection{Textual-Dominant Items}
\label{app:case_textual_dominant}

\begin{figure}[t]
    \centering
    \includegraphics[width=\linewidth]{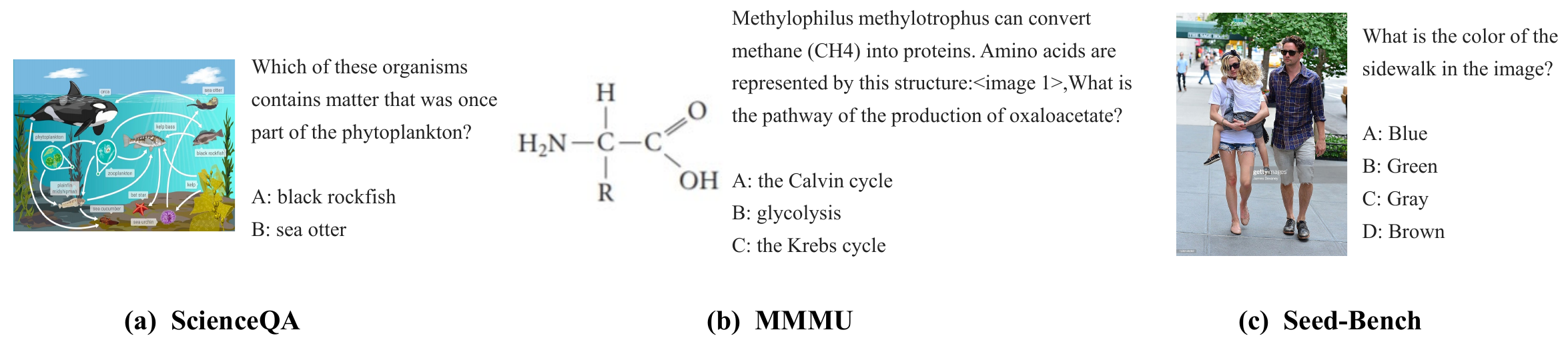}
    \caption{
    Examples of textual-dominant items from ScienceQA, MMMU, and SEED-Bench. 
    }
    \label{fig:case_textual_dominant}
\end{figure}

Figure~\ref{fig:case_textual_dominant} presents examples labeled as \textit{Textual-dominant}. 
In our annotation protocol, a textual-dominant item is one for which the textual-conditioned view is sufficient for human annotators to select the correct answer, while the visual-conditioned view is not. 
This means that the original question and answer options provide enough semantic, factual, or commonsense constraints to recover the answer, whereas the image-option view alone does not specify the intended query. 
Textual dominance therefore often reveals cases where the benchmark item can be solved primarily from question wording, domain knowledge, or answer-option priors, rather than from multimodal integration.

\myparagraph{ScienceQA}
The ScienceQA example asks which organism contains matter that was once part of phytoplankton, with options black rockfish and sea otter. 
The textual-conditioned view is sufficient because the question explicitly specifies the relevant ecological concept: matter transfer from phytoplankton through a food chain. 
Given the options, a human annotator can use basic biology knowledge to infer that black rockfish is more directly associated with a phytoplankton-supported aquatic food chain. 
The text therefore provides a clear reasoning target and enough semantic constraints to select the correct answer.

The visual-conditioned view is less reliable. 
Although the image shows a food-web diagram containing several organisms and energy-transfer arrows, the original question is not available in this view. 
With only the image and the options, the annotator does not know which relation in the food web is being queried: the item could ask about predation, energy transfer, trophic level, producer-consumer relations, or matter flow from a particular organism. 
Because multiple plausible questions are compatible with the same image and options, the visual-conditioned view does not provide a sufficiently well-defined decision criterion. 
This makes the item textual-dominant: the question text determines the relevant ecological relation, while the image alone does not identify what should be answered.

\myparagraph{MMMU}
The MMMU example asks for the pathway involved in the production of oxaloacetate, with options including the Calvin cycle, glycolysis, and the Krebs cycle. 
The textual-conditioned view is sufficient because the key term ``oxaloacetate'' directly invokes biochemical knowledge. 
Oxaloacetate is a central intermediate in the Krebs cycle, so the correct answer can be inferred from the question and options without relying on the image. 
The surrounding context about methane conversion and amino acids further frames the question as a biochemical pathway problem.

In contrast, the visual-conditioned view provides only a generic amino acid structure and the answer options. 
This structure indicates the general form of amino acids, but it does not specify oxaloacetate or the metabolic pathway being queried. 
Without the textual cue, the options Calvin cycle, glycolysis, and Krebs cycle are all plausible biochemical processes, and the image does not determine which pathway should be selected. 
Thus, the answer is driven primarily by textual biochemical knowledge rather than by visual evidence.

\myparagraph{SEED-Bench}
The SEED-Bench example asks for the color of the sidewalk in the image, with options blue, green, gray, and brown. 
The textual-conditioned view is sufficient because the question explicitly names the target object, ``sidewalk'', and the answer options are colors. 
Even without the image, common visual priors make gray the most plausible answer, since sidewalks are typically gray. 
The text and options therefore create a strong answer prior.

The visual-conditioned view, however, is ambiguous. 
The image contains several visible colors: green vegetation, gray pavement, blue clothing, and other background colors. 
Because the original question is absent, the annotator does not know which object or region should be inspected. 
The correct option is not determined by the image-option view alone; it becomes clear only when the text specifies that the target is the sidewalk. 
This item is therefore textual-dominant because the decisive cue is the textual specification of the queried object, while the image without the question leaves the selection underdetermined.

\myparagraph{Summary}
These examples show that textual-dominant items may appear multimodal in format, but their answers can be recovered primarily from the question and answer options. 
Such items are not necessarily invalid for all purposes, but they provide limited evidence that a model must use visual information. 
They are especially important for benchmark diagnosis because they can inflate multimodal evaluation scores through textual shortcuts, domain priors, or option-driven reasoning.

\subsection{Visual-Dominant Items}
\label{app:case_visual_dominant}

\begin{figure}[t]
    \centering
    \includegraphics[width=\linewidth]{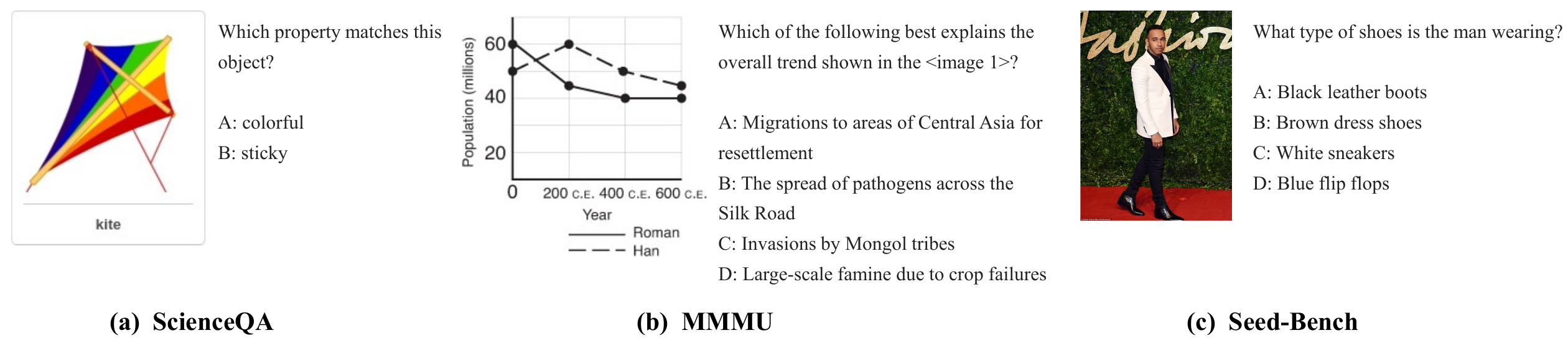}
    \caption{
    Examples of visual-dominant items from ScienceQA, MMMU, and SEED-Bench. 
    }
    \label{fig:case_visual_dominant}
\end{figure}

Figure~\ref{fig:case_visual_dominant} presents examples labeled as \textit{Visual-dominant}. 
In our annotation protocol, a visual-dominant item is one for which the visual-conditioned view is sufficient for human annotators to select the correct answer, while the textual-conditioned view is not. 
The visual-conditioned view contains the image and answer options, whereas the textual-conditioned view contains the original question and answer options. 
Thus, visual dominance indicates that the decisive evidence comes from the image-option view: without the image, the question either lacks the referenced object, lacks the visual trend, or lacks the visual attribute needed to identify the answer.

\myparagraph{ScienceQA}
The ScienceQA example asks which property matches the object, with options \textit{colorful} and \textit{sticky}. 
The visual-conditioned view is sufficient because the image clearly depicts a kite with multiple bright colors. 
Given the answer options, the visible appearance of the object directly supports \textit{colorful} as the correct choice. 
The image also contains the object label ``kite'', which further makes the visual view self-contained for this simple property-matching question.

The textual-conditioned view is insufficient because the question only refers to ``this object'' without showing the object. 
From the text and options alone, annotators know that they must choose a property, but they do not know what object is being described. 
Both options are generic properties that could apply to different objects in different contexts. 
Therefore, the correct answer cannot be reliably determined without the visual input. 
This item is visual-dominant because the image identifies the object and its salient property, while the text alone does not specify the target.

\myparagraph{MMMU}
The MMMU example asks which option best explains the overall trend shown in a population chart. 
The visual-conditioned view is sufficient because the chart shows the population trajectories of the Roman and Han populations over time. 
Both series exhibit an overall decline, and the answer options provide candidate historical explanations. 
By inspecting the plotted trend together with the options, annotators can infer that the question is asking for an explanation of the broad decline and select the option most consistent with that historical pattern.

The textual-conditioned view is not sufficient because the question depends on the phrase ``the overall trend shown in the image''. 
Without the chart, annotators cannot observe what the trend is: it could be an increase, a decline, a divergence between the two populations, or some other pattern. 
Although the options mention possible historical causes, the text alone does not specify which empirical pattern needs to be explained. 
As a result, the answer cannot be grounded from the question and options alone. 
This item is therefore visual-dominant because the decisive information is the visual trend in the chart.

\myparagraph{SEED-Bench}
The SEED-Bench example asks what type of shoes the man is wearing, with options including black leather boots, brown dress shoes, white sneakers, and blue flip flops. 
The visual-conditioned view is sufficient because the image clearly shows the man and his footwear. 
Given the options, the shoes can be identified from the visual appearance as black leather boots. 
The answer is determined by a local visual attribute of the person in the image.

The textual-conditioned view is insufficient because the question asks about an attribute of ``the man'' but provides no visual evidence about his appearance. 
The answer options list plausible footwear types, and none can be selected reliably from the wording alone. 
Any choice would amount to guessing or relying on an unsupported prior. 
Thus, the item is visual-dominant: the image provides the necessary attribute evidence, while the text only specifies what kind of visual attribute should be inspected.

\myparagraph{Summary}
These examples illustrate a common form of visual dominance in multimodal benchmarks. 
The question text may define the type of query, but the answer is determined by visual evidence such as object appearance, chart structure, or local attributes in the image. 
Without the image, the item is under-specified; with the image and answer options, the correct answer becomes identifiable.

\subsection{Synergistic Items}
\label{app:case_synergistic}

\begin{figure}[t]
    \centering
    \includegraphics[width=\linewidth]{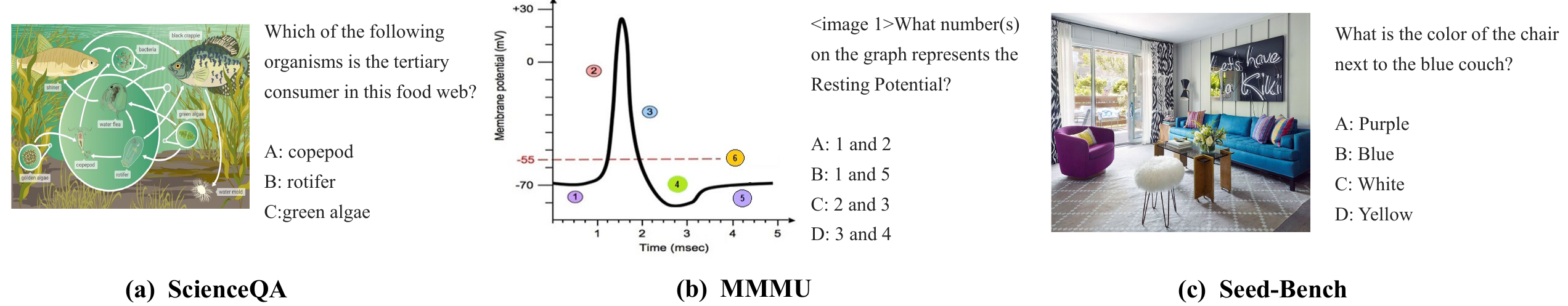}
    \caption{
    Examples of synergistic items from ScienceQA, MMMU, and SEED-Bench. 
    }
    \label{fig:case_synergistic}
\end{figure}

Figure~\ref{fig:case_synergistic} presents examples labeled as \textit{Synergistic}. 
In our annotation protocol, a synergistic item is one for which neither the visual-conditioned view nor the textual-conditioned view is sufficient for human annotators to select the correct answer, while the original full multimodal item remains solvable. 
Intuitively, synergistic items require combining the semantic target specified by the question with the visual evidence provided by the image. 
The text tells the annotator what relation, attribute, or concept should be identified, while the image provides the concrete objects, structures, or values needed to ground that query. 
When either side is removed, the answer becomes underdetermined.

\myparagraph{ScienceQA}
The ScienceQA example asks which organism is the tertiary consumer in a food web. 
The textual-conditioned view is insufficient because, although the question specifies the ecological concept of interest, it does not provide the food-web structure needed to determine which organism occupies the relevant trophic role. 
From the options alone, an annotator cannot infer which organism is connected to which others or how energy flows through the system.

The visual-conditioned view is also insufficient. 
The food-web diagram contains multiple organisms and feeding relations, but without the original question, the annotator does not know what aspect of the diagram should be identified. 
The same image could support many different questions, such as asking for a producer, a primary consumer, a predator, or an organism that eats a specific prey. 
Thus, the image and options alone do not specify that the target concept is ``tertiary consumer.''

The item becomes answerable only when the two modalities are combined. 
The question contributes the reasoning target---the trophic role to be identified---and the image provides the food-web structure needed to trace the corresponding organism. 
This is a characteristic synergistic case: the answer depends on combining conceptual information from the text with relational information from the image.

\myparagraph{MMMU}
The MMMU example asks which numbered position on an action-potential graph represents the resting potential. 
The textual-conditioned view is insufficient because the concept ``resting potential'' is defined relative to the visual graph and its numbered markers. 
Without the graph, the annotator cannot determine which numbered positions correspond to baseline membrane potential, depolarization, repolarization, or hyperpolarization.

The visual-conditioned view is likewise insufficient. 
The graph shows a membrane-potential curve with numbered markers, but without the original question, the annotator does not know which feature of the curve is being queried. 
The item could ask about resting potential, threshold, peak depolarization, repolarization, or after-hyperpolarization. 
Although the graph contains all of this information, the intended target cannot be identified from the image-option view alone.

Only the joint multimodal view resolves the task. 
The text specifies the electrophysiological concept to locate, while the image provides the graph and numbered positions that instantiate that concept. 
Neither modality alone is enough, but their combination yields a well-defined and answerable question. 
This is precisely the pattern captured by the synergistic label.

\myparagraph{SEED-Bench}
The SEED-Bench example asks for the color of the chair next to the blue couch. 
The textual-conditioned view is insufficient because the question specifies a target object ``the chair next to the blue couch'' relationally, but without the image, the annotator cannot see the room layout or identify the color of that chair. 
Although the answer options list candidate colors, the text alone does not provide enough information to determine which one is correct.

The visual-conditioned view is also insufficient. 
The room image contains several colored objects, including a blue couch, a purple chair, yellow accents, and other furniture. 
However, without the original question, the annotator does not know which object attribute is being queried. 
The image alone does not indicate whether the task concerns the couch, the chair, the walls, or some other object, nor does it specify that the relevant chair is the one adjacent to the blue couch.

The full multimodal item resolves this ambiguity. 
The text identifies the target through a relational description, and the image grounds that description by showing which chair is next to the blue couch and what color it is. 
This item is therefore synergistic: the answer depends on combining the textual reference with the visual scene.

\myparagraph{Summary}
These examples illustrate the core diagnostic meaning of synergy in multimodal benchmarks. 
A synergistic item is not merely one that contains both image and text, but one in which the two modalities make complementary contributions that are both necessary for answering the question. 
Such items provide the strongest evidence that a benchmark is actually testing multimodal understanding rather than unimodal shortcuts or answer-option priors.

\section{Experiment Extension}

\subsection{Statistical Significance Analysis}
\label{app:significance_analysis}

We further conduct paired statistical significance analysis on the human-labeled diagnostic validation set. 
Since all diagnosis methods are evaluated on the same set of items, we compare \ourq with the strongest baseline, M3IRT, using paired item-level tests. 
For both accuracy and macro-F1, we report the paired performance difference, 95\% confidence intervals estimated by item-level bootstrap resampling, and paired approximate-randomization \(p\)-values. 
For accuracy, we additionally report McNemar's test, which directly compares the number of items correctly classified only by \ourq and only by M3IRT.

Table~\ref{tab:significance_analysis} shows that the improvements of \ourq over M3IRT are statistically significant across all three benchmarks. 
On ScienceQA, \ourq improves accuracy by \(+0.037\) with a 95\% confidence interval of \([0.015, 0.060]\), and improves macro-F1 by \(+0.049\) with a 95\% confidence interval of \([0.026, 0.075]\). 
On MMMU and SEED-Bench, the gains are larger: \ourq improves accuracy by \(+0.061\) and \(+0.067\), respectively, and improves macro-F1 by \(+0.054\) and \(+0.070\), respectively. 
All bootstrap confidence intervals are strictly positive, and all approximate-randomization tests yield \(p<0.001\).

The McNemar results provide a complementary paired comparison for accuracy. 
For each benchmark, the number of items correctly classified only by \ourq is substantially larger than the number correctly classified only by M3IRT. 
For example, on MMMU, \ourq uniquely corrects 129 items, compared with 68 items uniquely corrected by M3IRT; on SEED-Bench, the corresponding counts are 115 and 48. 
The McNemar \(p\)-values are below \(0.01\) for all benchmarks. 
Aggregating across the validation items, \ourq improves accuracy by \(+0.055\) and macro-F1 by \(+0.058\), with both confidence intervals remaining strictly positive and all paired significance tests yielding \(p<0.001\). 
These results indicate that the observed gains of \ourq are unlikely to be explained by item-level sampling variation.

\begin{table}[t]
\centering
\caption{
Paired statistical significance analysis of \ourq over M3IRT on the human-labeled diagnostic validation set.
We report paired performance differences, 95\% bootstrap confidence intervals, approximate-randomization \(p\)-values, and McNemar's test for accuracy.
}
\label{tab:significance_analysis}
\resizebox{\linewidth}{!}{
\begin{tabular}{lccccccc}
\toprule
\multirow{2}{*}{Dataset}
& \multicolumn{3}{c}{Accuracy}
& \multicolumn{3}{c}{Macro-F1}
& \multirow{2}{*}{McNemar} \\
\cmidrule(lr){2-4}
\cmidrule(lr){5-7}
& \(\Delta\)
& 95\% CI
& \(p_{\mathrm{rand}}\)
& \(\Delta\)
& 95\% CI
& \(p_{\mathrm{rand}}\)
&  \\
\midrule
ScienceQA
& +0.037
& [0.015, 0.060]
& \(<0.001\)
& +0.049
& [0.026, 0.075]
& \(<0.001\)
& \(87/50,\ p=0.002\) \\

MMMU
& +0.061
& [0.035, 0.089]
& \(<0.001\)
& +0.054
& [0.025, 0.079]
& \(<0.001\)
& \(129/68,\ p<0.001\) \\

SEED-Bench
& +0.067
& [0.042, 0.091]
& \(<0.001\)
& +0.070
& [0.043, 0.095]
& \(<0.001\)
& \(115/48,\ p<0.001\) \\

\midrule
Average
& +0.055
& [0.040, 0.069]
& \(<0.001\)
& +0.058
& [0.044, 0.072]
& \(<0.001\)
& \(331/166,\ p<0.001\) \\
\bottomrule
\end{tabular}
}
\end{table}

\subsection{Comparison with Simple Fusion}
\label{app:simple_fusion}

To examine whether the improvement of \ourq can be attributed to a simple combination of intrinsic structural information and extrinsic model responses, we compare \ourq with a post-hoc profile fusion baseline. Unlike \ourq, which jointly calibrates structural priors and response-based item parameters, this baseline independently obtains the two diagnostic profiles and combines them using a fixed global weight.

Specifically, the fused diagnostic profile for item $j$ is defined as
\begin{equation}
\mathbf{q}^{\mathrm{fuse}}_j
=
\alpha \mathbf{q}^{\mathrm{str}}_j
+
(1-\alpha)\mathbf{q}^{\mathrm{resp}}_j,
\end{equation}
where $\mathbf{q}^{\mathrm{str}}_j$ is the normalized four-factor structural profile derived from LSMI, and $\mathbf{q}^{\mathrm{resp}}_j$ is the normalized response-based profile produced by M3IRT. The coefficient $\alpha$ controls the relative contribution of structural information. We evaluate three representative settings, $\alpha \in \{0.25, 0.50, 0.75\}$, and determine the dominant diagnostic factor from the resulting fused profile.

\begin{table}[t]
\centering
\caption{Comparison with post-hoc profile fusion baselines. We report accuracy and macro-F1 on the human-labeled validation set.}
\label{tab:simple_fusion}
\small
\setlength{\tabcolsep}{4pt}
\begin{tabular}{lcccccccc}
\toprule
\multirow{2}{*}{Method}
& \multicolumn{2}{c}{ScienceQA}
& \multicolumn{2}{c}{MMMU}
& \multicolumn{2}{c}{SEED-Bench}
& \multicolumn{2}{c}{Average} \\
\cmidrule(lr){2-3}
\cmidrule(lr){4-5}
\cmidrule(lr){6-7}
\cmidrule(lr){8-9}
& Acc. & Macro-F1
& Acc. & Macro-F1
& Acc. & Macro-F1
& Acc. & Macro-F1 \\
\midrule
$\alpha=0.25$
& 0.52 & 0.44 & 0.46 & 0.46
& 0.61 & 0.60 & 0.53 & 0.50 \\
$\alpha=0.50$
& 0.47 & 0.36 & 0.45 & 0.44
& 0.60 & 0.59 & 0.51 & 0.46 \\
$\alpha=0.75$
& 0.44 & 0.30 & 0.45 & 0.44
& 0.60 & 0.58 & 0.50 & 0.44 \\
\midrule
\textbf{\ourq}
& \textbf{0.61} & \textbf{0.59}
& \textbf{0.55} & \textbf{0.54}
& \textbf{0.69} & \textbf{0.67}
& \textbf{0.61} & \textbf{0.60} \\
\bottomrule
\end{tabular}
\end{table}

Table~\ref{tab:simple_fusion} shows that \ourq consistently outperforms all three fusion settings across the benchmarks. Even the strongest fusion baseline, with $\alpha=0.25$, achieves an average accuracy of 0.53 and macro-F1 of 0.50, compared with 0.61 and 0.60 for \ourq, respectively.

These results indicate that the improvement of \ourq cannot be reproduced by the tested fixed-weight fusion strategies. By incorporating structural information into item-parameter estimation and profile-level calibration, \ourq allows intrinsic structural evidence to interact with observed model responses during optimization. This provides a more effective integration of the two diagnostic signals than combining independently estimated profiles.

\begin{figure}[!t]
    \centering
    \includegraphics[width=\linewidth]{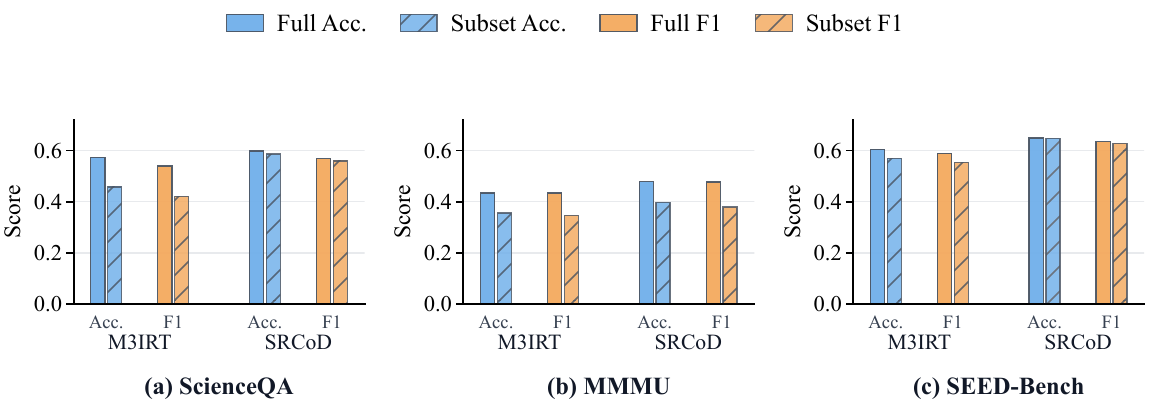}
    \caption{
    Diagnostic performance under a small-scale model pool. 
    We compare M3IRT and \ourq using responses from 3B-12B MLLMs and report accuracy and macro-F1. 
    \ourq consistently outperforms M3IRT and shows smaller degradation on the subset, indicating stronger robustness when response-side evidence is obtained from smaller and noisier models.
    }
    \label{fig:small_model_pool}
\end{figure}

\subsection{Robustness under a Small-Scale Model Pool}
\label{sec:small_model_pool}

To evaluate whether \ourq remains reliable when response-side evidence is limited to smaller models, we construct a small-scale MLLM pool with comparable parameter sizes, ranging from 3B to 12B parameters. This setting is more challenging than using a diverse pool of strong MLLMs, since smaller models often produce noisier and more biased modality-conditioned responses. We fit both M3IRT and \ourq using responses from this small-scale model pool and evaluate their diagnostic performance on the human-labeled validation set.

Figure~\ref{fig:small_model_pool} shows that \ourq remains consistently stronger than M3IRT when both methods are fitted using only responses from the small-scale model pool. On the full validation set, \ourq outperforms M3IRT across ScienceQA, MMMU, and SEED-Bench in both accuracy and macro-F1. The advantage is more pronounced on the subset evaluation. For example, on ScienceQA, M3IRT shows a substantial drop from the full set to the subset, whereas \ourq maintains nearly the same performance. A similar pattern appears on SEED-Bench, where \ourq remains stable while M3IRT degrades. On MMMU, both methods drop on the subset, but \ourq still achieves higher accuracy and macro-F1.

These results suggest that \ourq is less dependent on the reliability of response-side evidence alone. Since M3IRT relies purely on model response patterns, its estimates become less stable when the model pool consists of smaller and noisier MLLMs. \ourq mitigates this issue by anchoring item parameters with intrinsic item structure, leading to more robust item-level diagnosis under limited-model settings.

\begin{table}[t]
\centering
\caption{Behavioral validation on held-out model families. Values are absolute accuracy margins computed within the corresponding SRCoD-predicted diagnostic groups. Positive values indicate agreement with the expected behavioral patterns.}
\label{tab:heldout_families}
\setlength{\tabcolsep}{5pt}
\begin{tabular}{llcccc}
\toprule
Held-out Family & Benchmark
& $G_{\mathrm{vis}}$
& $G_{\mathrm{text}}$
& $G_{\mathrm{red}}$
& $G_{\mathrm{syn}}$ \\
\midrule
Qwen & SEED-Bench
& +0.28 & +0.11 & +0.26 & +0.15 \\
Qwen & ScienceQA
& +0.37 & +0.40 & +0.37 & +0.11 \\
Ministral & SEED-Bench
& +0.44 & +0.20 & +0.18 & +0.23 \\
Ministral & ScienceQA
& +0.30 & +0.35 & +0.27 & +0.13 \\
\bottomrule
\end{tabular}
\end{table}

\subsection{Behavioral Validation on Held-Out Model Families}
\label{app:heldout_families}

The model-pool subsampling experiment in Section~\ref{sec:model_pool_subsampling} examines the stability of \ourq under reduced diagnostic model pools. We further investigate whether the predicted modality-dependence profiles exhibit their expected behavioral patterns on model families that are not involved in \ourq fitting.

\myparagraph{Experimental Protocol}
We conduct a leave-one-model-family-out evaluation on ScienceQA and SEED-Bench, using Qwen and Ministral as two held-out model families. For each benchmark, we construct two family-level holdout settings. In the Qwen setting, all responses from Qwen-family models are excluded from \ourq fitting. Similarly, all Ministral-family responses are excluded in the Ministral setting.

\ourq is fitted exclusively using responses from the remaining model families. Each item is then assigned to the diagnostic category corresponding to its highest calibrated profile score. Importantly, responses from the held-out family are not used for either model fitting or item-profile assignment. They are used only for subsequent behavioral evaluation.

\myparagraph{Behavioral Evaluation}
We evaluate each held-out model family under four modality conditions: no modality ($0$), image only ($v$), text only ($t$), and image-text ($vt$). For each item and condition, correctness is first averaged across models within the held-out family. We then aggregate these item-level accuracies within each SRCoD-predicted diagnostic category.

For a given predicted category, let $A_c$ denote the mean held-out-family accuracy under modality condition $c$. We define four behavioral margins that characterize the expected response pattern of each diagnostic category:
\begin{equation}
\begin{aligned}
G_{\mathrm{vis}}
&= A_v-\max(A_t,A_0),\\
G_{\mathrm{text}}
&= A_t-\max(A_v,A_0),\\
G_{\mathrm{red}}
&= \min(A_v-A_0,A_t-A_0),\\
G_{\mathrm{syn}}
&= A_{vt}-\max(A_v,A_t).
\end{aligned}
\label{eq:heldout_margins}
\end{equation}

Each margin is computed on the item group assigned to its corresponding diagnostic category. A positive margin indicates that the held-out family exhibits the expected behavioral direction. In particular, $G_{\mathrm{vis}}>0$ and $G_{\mathrm{text}}>0$ indicate that the corresponding unimodal condition outperforms the alternatives. Similarly, $G_{\mathrm{red}}>0$ requires both unimodal conditions to outperform the no-modality baseline, while $G_{\mathrm{syn}}>0$ indicates an advantage of joint image-text input over either unimodal condition.

\myparagraph{Results}
Table~\ref{tab:heldout_families} shows that all 16 behavioral margins are positive across the two held-out model families and two benchmarks. In particular, items diagnosed as Synergistic consistently exhibit higher average accuracy under joint image-text input than under either unimodal condition. The Visual-dominant, Textual-dominant, and Redundant groups likewise exhibit their expected modality-specific behavioral patterns.

These results indicate that the diagnostic categories predicted by \ourq preserve their expected aggregate behavioral characteristics on unseen model families. This provides further evidence that structure-response co-calibration captures modality-dependence patterns that generalize beyond the diagnostic model pool used during fitting.






\end{document}